\documentclass{article}

\usepackage{microtype}
\usepackage{graphicx}
\usepackage{booktabs} 

\usepackage{stmaryrd}
\usepackage{threeparttable,color}
\usepackage{balance}
\usepackage{bm}
\usepackage{subfig}
\usepackage{graphicx}
\usepackage{amsmath,amssymb,amsfonts}
\usepackage{booktabs}
\usepackage{multirow}

\usepackage{hyperref}

\usepackage[accepted]{icml2021}

\icmltitlerunning{Graph Feature Gating Networks}

\begin{document}

\twocolumn[
\icmltitle{Graph Feature Gating Networks}



\icmlsetsymbol{equal}{*}

\begin{icmlauthorlist}
\icmlauthor{Wei Jin}{to}
\icmlauthor{Xiaorui Liu}{to} 
\icmlauthor{Yao Ma }{to}
\icmlauthor{Tyler Derr}{goo}
\icmlauthor{Charu Aggarwal}{ed}
\icmlauthor{Jiliang Tang}{to}
\end{icmlauthorlist}

\icmlaffiliation{to}{Department of Computer Science, Michigan State University, Michigan, USA}
\icmlaffiliation{goo}{Department of Electrical Engineering and Computer Science, Vanderbilt University, Tennessee, USA}
\icmlaffiliation{ed}{IBM T. J. Watson Research Center, New York, USA}

\icmlcorrespondingauthor{Wei Jin}{jinwei2@msu.edu }


\icmlkeywords{Machine Learning, ICML}

\vskip 0.3in
]



\printAffiliationsAndNotice{}  

\begin{abstract}
Graph neural networks (GNNs) have received tremendous attention due to their power in learning effective representations for graphs. Most GNNs follow a message-passing scheme where the node representations are updated by aggregating and transforming the information from the neighborhood. Meanwhile, they adopt the same strategy in aggregating the information from different feature dimensions. However, suggested by social dimension theory and spectral embedding, there are potential benefits to treat the dimensions differently during the aggregation process. In this work, we investigate to enable heterogeneous contributions of feature dimensions in GNNs. In particular, we propose a general graph feature gating network (GFGN) based on the graph signal denoising problem and then correspondingly introduce three graph filters under GFGN to allow different levels of contributions from feature dimensions. Extensive experiments on various real-world datasets demonstrate the effectiveness and robustness of the proposed frameworks.  
\end{abstract}

\section{Introduction}
Many types of real-world data can be naturally denoted as graphs such as social networks, transportation networks, and biological networks~\cite{battaglia2018relational-survey,wu2019comprehensive-survey,zhou2018graph-survey}.  Recent years have witnessed great success from graph neural networks (GNNs) in extracting effective graph representations and tackling graph-related applications~\cite{ma2020deep,wu2019comprehensive-survey,zhou2018graph-survey}. GNNs adopt a message-passing scheme, where for each given node, its node representation is obtained by aggregating and transforming the representations of its neighbors~\cite{mpnn}.
Specifically, in most GNN layers, node $v_i$'s representation is updated through the following process,

\begin{small}
\vskip -3.5ex
\begin{align}
\mathbf{H}_{i} \leftarrow \operatorname{Transform}(\sum\limits_{j\in \mathcal{N}(v_i)\cup{\{v_i\}}} c_{ij}{\bf H}_j),
\label{eq:GNNaggregation}
\end{align}
\end{small}where ${\bf H}_i\in \mathbb{R}^{1\times d}$ is the representation of node $v_i$ with $d$ indicating its dimension, $\mathcal{N}(v_i)$ denotes the neighborhood of $v_i$, and $c_{ij}$ is the aggregation coefficient for the node pair $(v_i, v_j)$. Though the existing GNNs consider different contributions of neighbors via $c_{ij}$, they use the exact same strategy (coefficients) to aggregate information from each feature dimension (or each dimension of ${\bf H}_j$). This naturally raises the question: should we consider each feature dimension differently during the aggregation?

\begin{figure}[tb]%
     \centering
     \subfloat[Feature dimension A]{{\includegraphics[width=0.5\linewidth]{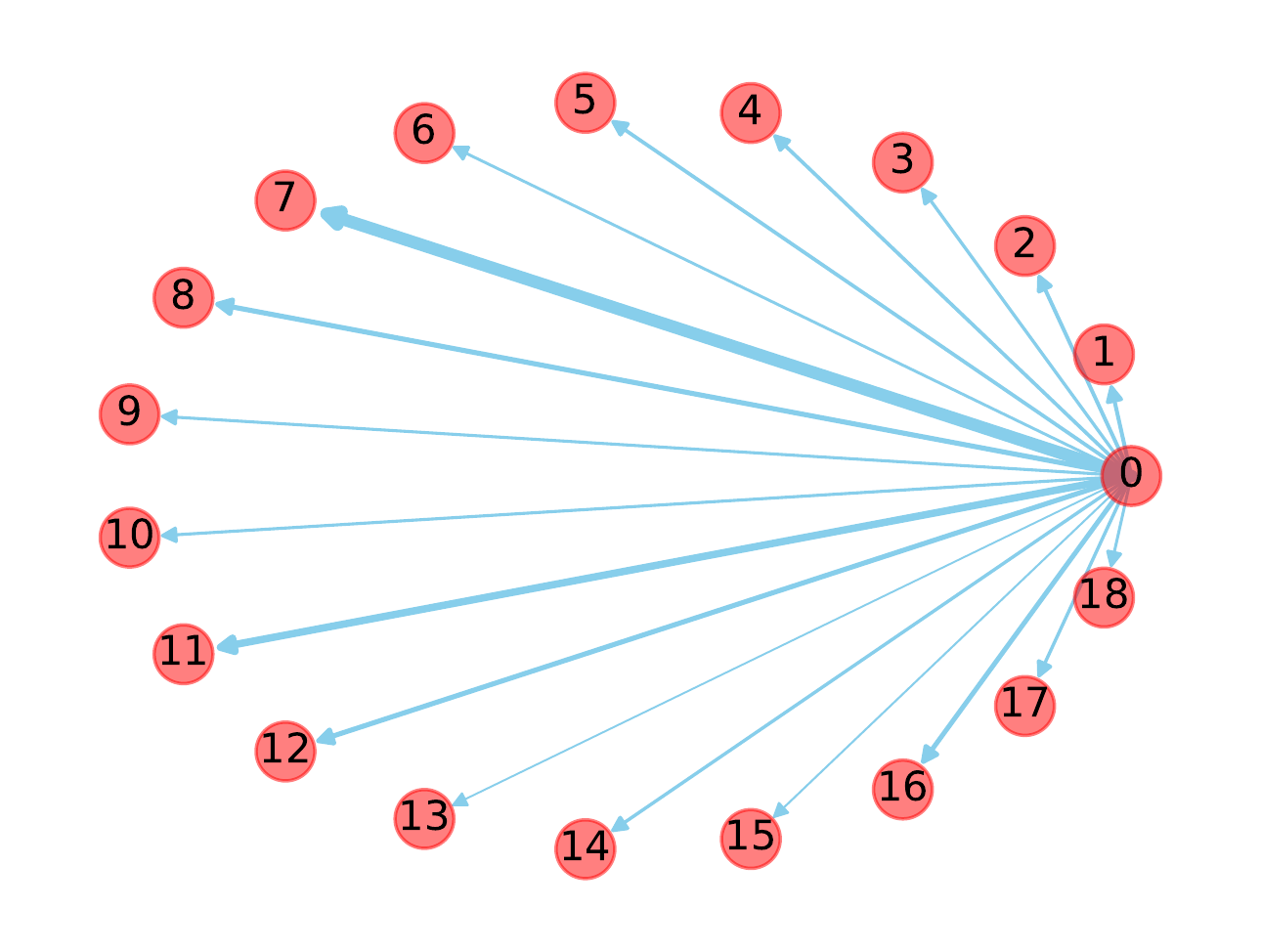} }}%
     \subfloat[Feature dimension B]{{\includegraphics[width=0.5\linewidth]{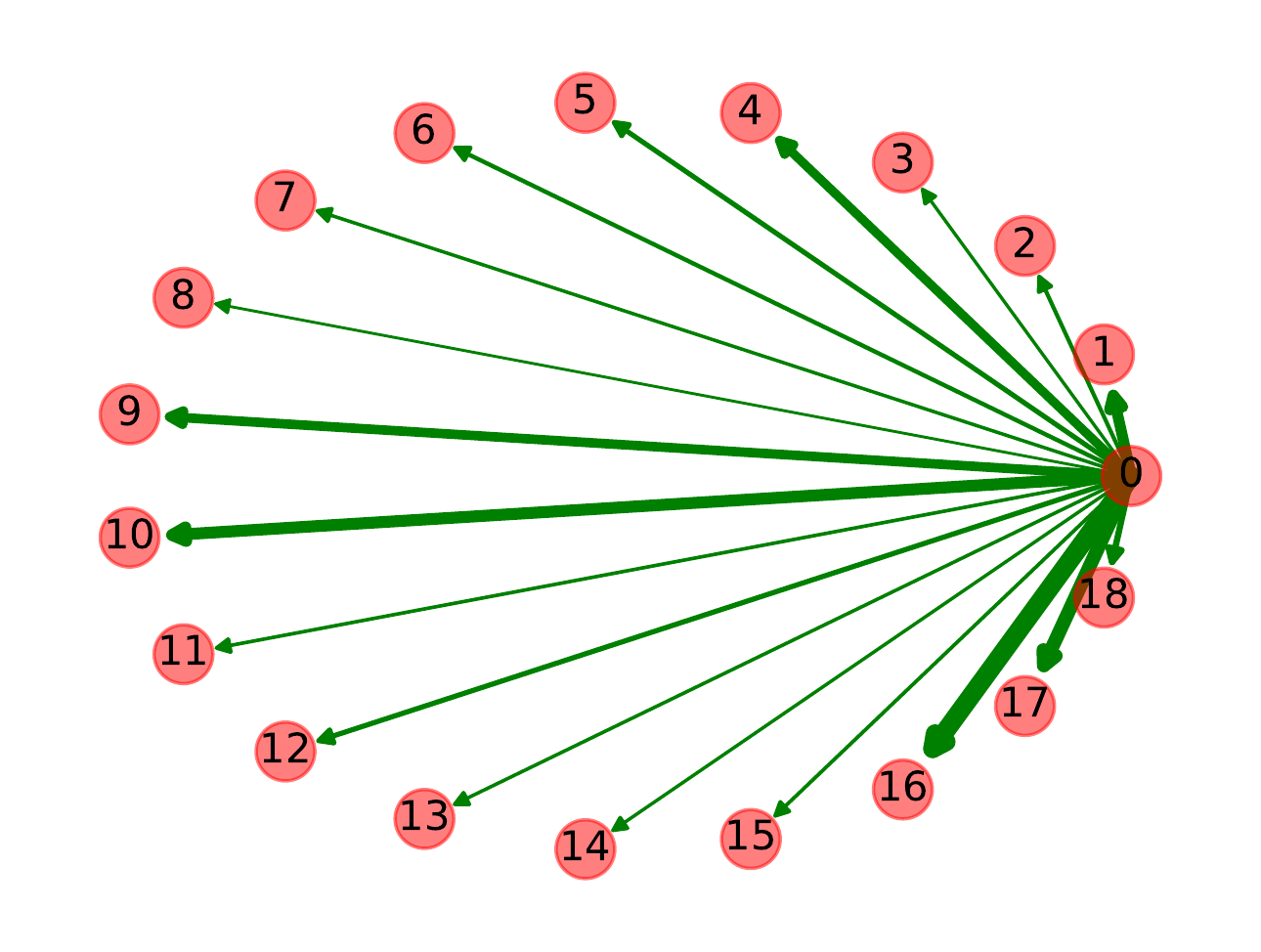} }}%
    \qquad
\vspace{-0.5em}
\caption{Feature similarities between user $0$ and its neighbors in two feature dimensions on the BlogCatalog dataset.}
\vspace{-1em}
\label{fig:dimensions}
\end{figure}

{\bf Motivation Example 1: Social Dimension Theory.} Social dimension theory indicates that the connections in social networks are not homogeneous since different people are connected due to diverse reasons~\cite{tang2009relational,tang2009scalable}. 
Figure~\ref{fig:dimensions} demonstrates the feature similarities between one user (denoted as $0$), and its  neighbors (indicated by $\{1,2,\cdots,18\}$), for two selected feature dimensions on the BlogCatalog dataset~\cite{huang2017label}\footnote{We perform PCA to reduce the dimension to 2. We also note that similar patterns are observed for the original features.}. The
width of arcs in these figures indicates their similarities: the wider the arc is, the more similar the connected nodes are. We can observe that user $0$ has very different similarities with its neighbors in different feature dimensions.  For example, user $0$ and user $7$ are highly similar in the dimension A while presenting a low similarity in the dimension B; user $0$ and user $10$ are very different in the dimension B but with a high similarity in the dimension A. In addition to social networks, social dimension theory can be applied to other types of graphs. For example, in a website graph, a website can connect to other websites (e.g., advertisement websites and donation websites) for different reasons (or different feature dimensions).
These phenomena reveal that nodes have heterogeneous connections with their neighbors in different feature dimensions. 

{\bf Motivation Example 2: Spectral Embedding.} By selecting top eigenvectors of the graph Laplacian matrix as the node representation, the method of spectral embedding embeds the graph into a low-dimensional space where each dimension corresponds to an eigenvector~\cite{von2007tutorial,ng2002spectral}. We show that, for a node $v_i$, different dimension of its spectral embedding have different similarities (smoothness) with its neighborhood, controlled by the eigenvalues. More details can be found from Appendix A.1 in the supplementary file. These differences across dimensions are overlooked by current GNN designs.

The aforementioned two motivation examples suggest that we should consider each feature dimension differently during the aggregation process. Thus, in this paper, we aim to design a new aggregation scheme which enables feature dimensions to contribute differently to the aggregation process. Inspired by the general framework recently introduced in~\cite{ma2020unified}, we introduce a general graph feature gating network (GFGN) based on the graph signal denosing problem. Based on GFGN, we introduce three graph filters that allow different-level contributions of feature dimensions. Our contributions can be summarized as follows:

(1) Based on the graph signal denoising problem, we design a general framework GFGN that allows different feature dimensions to contribute differently during aggregation.

(2) Under the general framework of GFGN, we propose three graph filters to capture different-level contributions of feature dimensions to meet different demands.

(3) Extensive experiments on various datasets validate the effectiveness and advantages of the proposed GFGN. We also show that GFGN is more robust to severe random noise.

\section{Preliminary}
In this section, we first introduce the notations used in this paper and then discuss some preliminaries of graph neural networks. We denote a graph as $\mathcal{G}=({\mathcal{V}},{\mathcal{E}}, {\bf X})$ where $\mathcal{V}=\{v_1, v_2, ..., v_N\}$ is the set of $N$ nodes, $\mathcal{E} \subseteq{\bf\mathcal{V}\times \bf\mathcal{V}}$ is the set of edges describing the relations between nodes, and ${\bf X} = [{\bf x}_1,{\bf x}_2,\ldots,{\bf x}_N]\in \mathbb{R}^{N\times D}$ indicates the node feature matrix with ${\bf x}_i$ representing the $D$-dimensional feature vector of node $v_i$. The relations between nodes can also be described by an adjacency matrix $\mathbf{A} \in \{0,1\}^{N \times N}$ where $\mathbf{A}_{ij}=1$ only when there exists an edge between nodes $v_i$ and $v_j$, otherwise $\mathbf{A}_{ij}=0$. Hence, we can also denote a graph as $\mathcal{G}=({\bf A},{\bf X})$. In the setting of semi-supervised node classification, a subset of nodes $\mathcal{V}_{L}\subset \mathcal{V}$ are associated with corresponding labels $\mathcal{Y}_L$. The goal of node classification is to infer the labels for the unlabeled data by learning a mapping function. Its objective can be summarized as follows,

\begin{small}
\vskip -3.5ex
\begin{equation}
\min _{\theta} \sum_{v_i \in \mathcal{V}_{L}} \ell\left(\hat{y}_{i}, y_{i}\right),
\end{equation}
\end{small}where $\hat{y}_{i}$ and $y_{i}$ denote the output of the mapping function and true label for node $v_i$ respectively, and $\ell(\cdot, \cdot)$ denotes the loss function. 

The mapping function is often implemented as a graph neural network. Most graph neural networks follow a message passing scheme where the node representation is obtained by aggregating the representation of its neighbors and updating its own representation. We use $\mathbf{H}^{(l)}$ to denote the node hidden representation matrix at the $l$-th layer. Further, the operation in the $l$-th graph neural network layer can be described as,

\begin{small}
\vskip -4.5ex
\begin{equation}
 \mathbf{H}_i^{(l)} =  \operatorname{Transform }\left(\operatorname{Aggregate}\left(\mathbf{H}_{j}^{(l-1)} \mid v_j \in \mathcal{N}(v_i)\cup{v_i}\right)\right),
\label{eq:message_pass}
\end{equation}
\end{small}where $\mathbf{H}^{(l)}_i$ denotes the $l$-th layer hidden representation of node $v_i$. Then the predicted probability distribution of a $k$-layer graph neural network can be formulated as,

\begin{small}
\vskip -2.5ex
\begin{equation}
\hat{y}_{v_i}=\operatorname{softmax}\left({\bf H}^{(k)}_i\right).
\end{equation}
\end{small}A representative example of graph neural network is Graph Convolution Network (GCN)~\cite{kipf2016semi}. It implements the message passing scheme in Eq.~\eqref{eq:message_pass} as 

\begin{small}
\vskip -2.5ex
\begin{equation}
\mathbf{H}^{(l)} = {\sigma} \left(\tilde{\bf D}^{-1 / 2}{\bf \tilde{A}} \tilde{\bf D}^{-1 / 2} {\mathbf H}^{(l-1)} {\mathbf W}^{(l)}\right),
\label{eq:gcn}
\end{equation}
\end{small}where ${\sigma}$ denotes the activation function such as ReLU, $\tilde{\bf A}={\bf A}+ {\bf I}$ and $\tilde{\bf D}$ is the diagonal matrix of $\tilde{\bf A}$ with $\tilde{\bf D}_{ii} = 1 + \sum_{j} {\bf A}_{ij}$ and ${\bf W}$ is the parameter matrix for feature transformation. 
In the following discussions, for simplicity, we will use $\bf{H}^\prime$ and $\bf{H}$ as the hidden representations of the current layer and previous layer, respectively.



\section{The Proposed Framework}
\label{sec:framework}

In this section, we design graph feature gating network (GFGN) inspired by the general framework recently introduced in~\cite{ma2020unified}. We first introduce the general GFGN framework. Then we propose three graph filters to allow different-level contributions of feature dimensions (or different-level feature smoothing) to the aggregation process. Finally we provide an analysis on GFGN. 

\subsection{GFGN via Graph Signal Denoising}
\label{sec:signal_denoise}

In~\cite{ma2020unified}, numerous existing GNNs have been proven to exactly (or approximately) solving the graph signal denoising problem. Given its generalization, we also propose to design the general GFGN framework based on the graph signal denoising problem. Assume that we are given a noisy graph signal ${\bf x} = {\bf f}^*+\eta$, where ${\bf f}^*$ is the clean graph signal and $\eta$ is uncorrelated Gaussian noise. Note that the graph signal is single-channel, i.e., ${\mathbf x} \in \mathbb {R}^{N\times 1}$, with the $i$-th element corresponding to node $v_i$. The goal of graph signal denoising is to recover ${\bf f}^*$ from ${\bf x}$. Since the clean signal ${\bf f}^*$ is often assumed to be smooth with respect to the underlying graph $\mathcal{G}$, we can adopt the following objective function to recover the clean signal: 

\begin{small}
\vskip -2.5ex
\begin{equation}
    \min_{\bf f}~~ g({\bf f}) = \|{\bf f}-{\bf x}\|^2 + c{\bf f}^\top {\bf L} {\bf f} ,
\label{eq:opti_signal}
\end{equation}
\end{small}where $c$ denotes the smoothing coefficient controlling how smooth we want the graph signal to be; and ${\bf L}$ is the Laplacian matrix of the underlying graph $\mathcal{G}$. If we adopt the normalized Laplacian matrix, i.e. ${\bf {L}} = {\bf I} -  {\bf { D}}^{-\frac{1}{2}}{\bf {A}}{\bf {D}}^{-\frac{1}{2}}$, Eq.~\eqref{eq:opti_signal} can be rewritten as,

\begin{small}
\vskip -4.5ex
\begin{equation}
    \min_{\bf f}~~ g({\bf f}) = \|{\bf f}-{\bf x}\|^2 + \frac{c}{2} \sum\limits_{v_i\in \mathcal{V}}\sum\limits_{v_j\in \mathcal{N}(v_i)}(\frac{{\bf f}_{i}}{\sqrt{d_{i}}}-\frac{{\bf f}_{j}}{\sqrt{d_{j}}})^{2},
\label{eq:normed_lap}
\end{equation}
\end{small}where $\mathcal{N}(v_i)$ denotes the neighbors of node $v_i$. It is clear that the second term in Eq.~\eqref{eq:normed_lap} is small when connected nodes share similar signals, which can indicate the smoothness of ${\bf f}$. 
We then consider the optimization process of Eq.~\eqref{eq:normed_lap}. For node $v_i$ with its corresponding signal value ${\bf f}_i$, the gradient of $g({\bf f})$ with respect to ${\bf f}_i$ is calculated as: 

\begin{small}
\vskip -4.5ex
\begin{align}
    \frac{\partial g({\bf f})}{\partial \mathbf{f}_{i}} & = 2({\bf f}_i-{\bf x}_i) + 2c\sum\limits_{v_j\in \mathcal{N}(v_i)}(\frac{{\bf f}_{i}}{{d_{i}}}-\frac{{\bf f}_{j}}{\sqrt{{d_i}d_{j}}})  \nonumber \\
    & = 2({\bf f}_i-{\bf x}_i) + 2c{\bf f}_i - 2c\sum\limits_{v_j\in \mathcal{N}(v_i)}\frac{{\bf f}_{j}}{\sqrt{{d_i}d_{j}}} .
\end{align}
\end{small}Then, by taking one step gradient descent at ${\bf x}_i$ with a step size $\epsilon$, we have:

\begin{small}
\vskip -4.5ex
\begin{align}
{\bf f}_i \leftarrow {\bf x}_i - \epsilon 
\frac{\partial g({\bf f})}{\partial \mathbf{f}_{i}} 
& = (1-2\epsilon c){\bf x}_i + \sum\limits_{v_j\in \mathcal{N}(v_i)}2\epsilon{c}\frac{{\bf x}_{j}}{\sqrt{{d_i}d_{j}}}. 
\label{eq:one_step_gradient}
\end{align}
\end{small}Note that the process in Eq.~\eqref{eq:one_step_gradient} resembles the aggregation process in graph neural networks. Meanwhile, $c$ in Eq.~\eqref{eq:one_step_gradient} naturally controls how the neighbors contribute to the aggregation process in that signal channel (or feature dimension). In other words, we can allow each dimension to contribute differently by varying the definitions of $c$. Therefore, the proposed general GFGN framework is to solve the following generalized graph signal denoising problem as:

\begin{small}
\vskip -4.5ex
\begin{equation}
    \min_{\bf f}~~ g({\bf f}) = \|{\bf f}-{\bf x}\|^2 + \Omega(\mathcal{C}) \sum\limits_{v_i\in \mathcal{V}}\sum\limits_{v_j\in \mathcal{N}(v_i)}(\frac{{\bf f}_{i}}{\sqrt{d_{i}}}-\frac{{\bf f}_{j}}{\sqrt{d_{j}}})^{2},
\label{eq:general_GFGN}
\end{equation}
\end{small}where $\Omega(\mathcal{C})$ is a variable (either predefined or learnable) that can be defined differently according to what feature smoothing we aim to achieve. In the following subsections, we introduce three ways to define $\Omega(\mathcal{C})$ which leads to three new graph filers: GFGN-graph, GFGN-neighbor and GFGN-pair. Note that we present the general GFGN framework in the form of a single channel signal (or feature dimension) in Eq.~(\ref{eq:general_GFGN}) to ease our discussion but we will discuss multi-channel signals (or multi-dimensional features) when we introduce the new filters.

\subsection{GFGN-graph: Graph-level GFGN}
\label{sec:global}
Before we introduce GFGN-graph, we use the GCN model to further demonstrate that GCN does not distinguish different contributions from feature dimensions. When we consider the activation function and multi-dimensional features, Eq.~\eqref{eq:one_step_gradient} can be rewritten as:

\begin{small}
\vskip -4.5ex
\begin{equation}
{\bf H}'_i = \sigma\left((1-2\epsilon c){\bf H}_i{\bf W} + \sum\limits_{v_j\in \mathcal{N}(v_i)}2\epsilon{c}\frac{{\bf H}_{j}{\bf W}}{\sqrt{{d_i}d_{j}}} \right), 
\label{eq:update_rule}
\end{equation}
\end{small}where ${\bf H}_i$ and ${\bf H}'_i$ denote the input and output multi-dimensional feature representations of node $v_i$, respectively; ${\bf W}$ is the parameter matrix for feature transformation; and $\sigma$ denotes the nonlinear activation function. If we set $\epsilon=\frac{1}{2c}$ and utilize ${\bf I} -  {\bf \tilde{ D}}^{-\frac{1}{2}}{\bf \tilde{A}}{\bf \tilde{D}}^{-\frac{1}{2}}$ as the Laplacian matrix in Eq.~\eqref{eq:opti_signal}, Eq.~\eqref{eq:update_rule} is equivalent to GCN in Eq.~\eqref{eq:gcn}. This observation suggests that GCN uses the same scalar $c$ for all feature dimensions. Therefore, we develop GFGN-graph which uses a vector ${\bf c}$ with the same size of the feature dimension with its $k$-th element ${\bf c}_k$ indicating the feature smoothness coefficient for the $k$-th feature dimension. Given that ${\bf c}$ is a vector potentially with a large number of dimensions and we do not have prior knowledge about the smoothness of each feature dimension, it is impractical to manually define them as hyper-parameters. Therefore, we propose to learn these coefficients in a data driven way, which leads to the following to update hidden representations,

\begin{small}
\vskip -4.5ex
\begin{equation}
    {\bf H}^\prime_i = \sigma\left(({\bf1}-{\bf s}) \odot {\bf H}_i{\bf W} + \sum\limits_{v_j\in \mathcal{N}(v_i)} {\bf s} \odot  \frac{{\bf H}_{j}{\bf W}}{\sqrt{{d_i}d_{j}}} \right),
\end{equation}
\end{small}where $\odot$ denotes the element-wise multiplication, ${\bf 1}$ is a row vector of $D$-dimension with all one entries, and ${\bf s}=2\epsilon{\bf c}$ is a row vector of dimension $D$. For the dimension $k$, all nodes in the graph share the same smoothing coefficient ${\bf s}_k$ Therefore, in our work, ${\bf s}$ is modeled to include the whole graph structure information as follows:

\begin{small}
\vskip -4.5ex
\begin{equation}
    {\bf s} =  \lambda \cdot\operatorname{sigmoid}\left(\text{Pool}\left({\bf H}_1{\bf W}, {\bf H}_2{\bf W},\dots,{\bf H}_N{\bf W}\right) {\bf W}_s\right),
\label{eq:model_s}
\end{equation}
\end{small}where $\text{Pool}(\cdot)$ denotes a pooling operation which takes a set of node representations as input and outputs a combined representation; $\text{Pool}\left({\bf H}_1{\bf W}, {\bf H}_2{\bf W},\dots,{\bf H}_N{\bf W}\right)$ can be viewed as the representation for the whole graph; ${\bf W}_s\in\mathbb{R}^{D\times{D}}$ is the parameter matrix for transforming the graph representation to the smoothness vector; $\operatorname{sigmoid}(\cdot)$ is the sigmoid function and $\lambda$ is a hyperparameter controlling the output range of the gating units. In this paper, we adopt mean pooling as the pooling operation. Furthermore, by making the smoothing score ${\bf s}$ learnable, we have the potential to enhance robustness of the model under random noise. In particular, when the connections in the graph are very noisy, the feature smoothness between connected nodes can be dramatically varied: some feature dimensions are more smooth than others. Under such scenario, the proposed aggregation scheme can adaptively learn to assign different smoothness coefficients with respect to each feature dimension and consequently improve the performance under random noise. More details can be found in Section~\ref{sec:random_noise}.

\subsection{GFGN-neighbor: Neighborhood-level GFGN}

GFGN-graph assumes the same smoothness for each feature dimension over all nodes. However, in reality, each node could have different levels of smoothness. Thus, we propose GFGN-neighbor that allows each node to have the unique smoothness with its neighbors with still assigning different coefficients over feature dimensions. Based on  Eq.~(\ref{eq:general_GFGN}), we assign  different smoothness coefficient $c_i$ for each node $v_i$ and the optimization problem can be written as:

\begin{small}
\vskip -4.5ex
\begin{equation}
    \min_{\bf f}~~ g({\bf f}) = \|{\bf f}-{\bf x}\|^2 + \frac{1}{2}\sum\limits_{v_i\in \mathcal{V}}c_i \cdot\sum\limits_{v_j\in \mathcal{N}(v_i)}(\frac{{\bf f}_{i}}{\sqrt{d_{i}}}-\frac{{\bf f}_{j}}{\sqrt{d_{j}}})^{2}. \nonumber
\end{equation}
\end{small}For node $v_i$ with its corresponding value ${\bf f}_i$, the gradient of $g({\bf f})$ with respect to ${\bf f}_i$ is calculated as:

\begin{small}
\vskip -4.5ex
\begin{align}
    \frac{\partial g({\bf f})}{\partial \mathbf{f}_{i}} & = 2({\bf f}_i-{\bf x}_i) + \sum\limits_{v_j\in \mathcal{N}(v_i)}(c_i+c_j)(\frac{{\bf f}_{i}}{{d_{i}}}-\frac{{\bf f}_{j}}{\sqrt{{d_i}d_{j}}}).  
\end{align}
\end{small}Then, the one step gradient descent at ${\bf x}_i$ with a step size $\epsilon$ can be expressed as follows,

\begin{small}
\vskip -4.5ex
\begin{align}
& {\bf f}_i \leftarrow  ~ {\bf x}_i - \epsilon \frac{\partial g({\bf f})_{|\bf f={\bf x}}}{\partial \mathbf{f}_{i}} \nonumber \\ 
& = (1-\sum\limits_{v_j\in \mathcal{N}(v_i)}\frac{\epsilon{(c_i+c_j)}}{d_i}){\bf x}_i + \sum\limits_{v_j\in \mathcal{N}(v_i)}\epsilon{(c_i+c_j)}\frac{{\bf x}_{j}}{\sqrt{{d_i}d_{j}}}.  \nonumber
\end{align}
\end{small}If we allow $\epsilon$ to be adaptive on each neighbor of $v_j$, denoted as $\epsilon_j$,  and set $\epsilon_j = \frac{s_i}{(c_i+c_j)}$, the above equation can be rewritten as:

\begin{small}
\vskip -4.5ex
\begin{align}
    {\bf f}_i \leftarrow (1-s_i) {\bf x}_i +  \sum\limits_{v_j\in {\mathcal{N}}(v_i)}s_i \frac{{\bf x}_{j}}{\sqrt{{d_i}d_{j}}}.  \label{eq:update_rule_neighbor}
\end{align}
\end{small}By rewriting Eq.~\eqref{eq:update_rule_neighbor} with multi-dimensional features, we obtain GFGN-neighbor as, 

\begin{small}
\vskip -4.5ex
\begin{align}
    {\bf H}^\prime_i = \sigma\left(({\bf 1}-{\bf s}_i) \odot {\bf H}_i{\bf W} +  \sum\limits_{v_j\in {\mathcal{N}}(v_i)}{\bf s}_i \odot \frac{{\bf H}_{j}{\bf W}}{\sqrt{{d_i}d_{j}}} \right),
\end{align}
\end{small}where ${\bf s}_i\in{\mathbb{R}^{1\times{D}}}$ is the smoothness vector for node $v_i$ with its neighbors over all feature dimensions. Directly parametrizing ${\bf s}_i (i=1,2,\cdots,N)$ will introduce a $N\times{D}$ matrix. Hence, to reduce the number of parameters, we model ${\bf s}_i$ by employing gating units as:

\begin{small}
\vskip -3.5ex
\begin{align}
{\bf s}_i =  \lambda \cdot\operatorname{sigmoid}\left(\left({\bf H}_i{\bf W}~\|~ {\text{Pool}\left({\bf H}_j{\bf W} \mid v_j\in {\mathcal{N}}(v_i)\right)}\right){\bf W_s} \right), \nonumber
\end{align}
\end{small}where ${\text{Pool}\left({\bf H}_j{\bf W} \mid  v_j\in {\mathcal{N}}(v_i)\right)}$ can be viewed as the representation for the neighborhood of $v_i$, and $\|$ is the concatenation operation. As we can see in the above equation, ${\bf s}_i$ is obtained by transforming the node representation of $v_i$ and the neighborhood representation of $v_i$. Similar to GFGN-graph, GFGN-neighbor also enjoys the robustness advantage.

\subsection{GFGN-pair: Pair-level GFGN}
In GFGN-pair, we aim to give different smoothness coefficients to each node pair $(v_i, v_j)$. Based on  Eq.~(\ref{eq:general_GFGN}), we assign a smoothness coefficient $c_{ij}$ to each connected node pair $(v_i,v_j)$ as:

\begin{small}
\vskip -4.5ex
\begin{equation*}
    \min_{\bf f}~~ g({\bf f}) = \|{\bf f}-{\bf x}\|^2 + \frac{1}{2}\sum\limits_{v_i\in \mathcal{V}} \sum\limits_{v_j\in \mathcal{N}(v_i)}c_{ij}(\frac{{\bf f}_{i}}{\sqrt{d_{i}}}-\frac{{\bf f}_{j}}{\sqrt{d_{j}}})^{2},
\label{eq:normed_lap_pair}
\end{equation*}
\end{small}After performing one-step gradient descent at ${\bf x}_i$ with a step size of $\epsilon$, we have:

\begin{small}
\vskip -3.5ex
\begin{align}
& {\bf f}_i \leftarrow  ~ {\bf x}_i - \epsilon \frac{\partial g({\bf f})_{|\bf f={\bf x}}}{\partial \mathbf{f}_{i}} \nonumber \\ 
& = (1-\sum\limits_{v_j\in \mathcal{N}(v_i)}\frac{\epsilon{(c_{ij}+c_{ji})}}{d_i}){\bf x}_i + \sum\limits_{v_j\in \mathcal{N}(v_i)}\epsilon{(c_{ij}+c_{ji})}\frac{{\bf x}_{j}}{\sqrt{{d_i}d_{j}}}.   \nonumber
\end{align}
\end{small}Furthermore, by setting $s_{ij}=\epsilon(c_{ij}+c_{ji})$, we have,
\begin{align}
{\bf f}_i \leftarrow (1-\sum\limits_{v_j\in \mathcal{N}(v_i)}\frac{{s}_{ij}}{d_i}){\bf x}_i + \sum\limits_{v_j\in \mathcal{N}(v_i)}s_{ij}\frac{{\bf x}_{j}}{\sqrt{{d_i}d_{j}}} . 
\label{eq:update_rule_pair}
\end{align}
Extending Eq.~\eqref{eq:update_rule_pair} with multi-dimensional features, we design GFGN-pair as:

\begin{small}
\vskip -4.5ex
\begin{align}
    {\bf H}^\prime_i = \sigma\left(({\bf 1}-\sum\limits_{v_j\in \mathcal{N}(v_i)}\frac{{\bf s}_{ij}}{d_i})\odot{\bf H}_i {\bf W}+ \sum\limits_{v_j\in {\mathcal{N}}(v_i)}{\bf s}_{ij} \odot \frac{{\bf H}_{j}{\bf W}}{\sqrt{{d_i}d_{j}}} \right)
\label{eq:update_pair}
\end{align}
\end{small}where  ${\bf s}_{ij}$ is the smoothness vector for node pair $(v_i, v_j)$ over all feature dimensions. Instead of directly paramterizing all ${\bf s}_{ij}$, we model ${\bf s}_{ij}$ as follows,

\begin{small}
\vskip -3.5ex
\begin{align}
{\bf s}_{ij} =  \lambda \cdot\operatorname{sigmoid}\left(\left({\bf H}_i{\bf W}~\|~{\bf H}_j{\bf W}\right){\bf W_s} \right).
\label{eq:score_pair}
\end{align}
\end{small}As we can see in Eq.~\eqref{eq:score_pair}, ${\bf s}_{ij}\in{\mathbb{R}^{1\times{D}}}$ is obtained by transforming the node representations of $v_i$ and  $v_j$. 
Note that GFGN-pair is different from graph attention network (GAT) in two aspects: (1) GFGN-pair learns different scores for each node pair while the scores are also different for each feature dimension; (2) GAT uses softmax function to normalize the attention scores but  GFGN-pair treats the scores as gating units to control the amount of information coming from different feature dimensions.

\subsection{Understanding From the Spectral Domain}
In this subsection, we aim to provide some analysis for the proposed framework from the spectral domain. We start from the simplest  single-channel graph signal as described in Section~\ref{sec:signal_denoise}. Specifically, we use a single-channel signal ${\bf h} \in \mathbb{R}^{N\times 1}$ to represent any column of the matrix ${\bf H}{\bf W}$. Then, the update rule in Eq.~\eqref{eq:one_step_gradient} for this single channel signal can be rewritten follows, 

\begin{small}
\vskip -4.5ex
\begin{align}
{\bf h}' & = \sigma\left((1-2\epsilon{c}){\bf h} + 2\epsilon{c} {\bf { D}}^{-\frac{1}{2}}{\bf {A}}{\bf {D}}^{-\frac{1}{2}} {\bf h}\right)  \nonumber \\
& = \sigma\left(({\bf I} - s {\bf L}) {\bf h}\right),
\end{align}
\end{small}where $s=2\epsilon{c}$ and ${\bf L}={\bf I} - {\bf { D}}^{-\frac{1}{2}}{\bf {A}}{\bf {D}}^{-\frac{1}{2}}$. If we set $\sigma(\cdot)$ to be identity function and apply $K$ graph filter layers, the graph filtering process will be,

\begin{small}
\vskip -4.5ex
\begin{align}
{\bf h}^\prime = ({\bf I} - s {\bf L})^{K} {\bf h}.
\label{eq:linear_multiple_layer}
\end{align}
\end{small}The Laplacian matrix ${\bf L}$ has the following eigen-decomposition ${\bf L}= {\bf U} {\bm \Lambda} {\bf U}^{\top}$, where ${\bf U}$ consists of all eigenvectors of ${\bf L}$ as columns and ${\bm \Lambda}=\operatorname{diag}(\lambda_1,\ldots,\lambda_N)$ is a diagonal matrix consisting of the corresponding eigenvalues in its diagonal. 
Note that ${\bf U}$ can be regarded as frequency components, and their corresponding eigenvalues measure their frequency~\cite{shuman2013emerging}. Given this decomposition, we can rewrite Eq.~\eqref{eq:linear_multiple_layer} as follows, 

\begin{small}
\vskip -2.5ex
\begin{equation}
    ({\bf I} - s {\bf L})^{K} {\bf h}  = {\bf U} {\bm \hat{\Lambda}}^{K} {\bf U}^{\top} {\bf h}, \label{eq:filter_process}
\end{equation}
\end{small}where ${\bm \hat{\Lambda}}^{K} =\operatorname{diag}\left((1-s\lambda_1)^K,\ldots,(1-s\lambda_N)^K\right)$ with the diagonal elements correspond to filter spectral coefficients. Specifically, the filter spectral coefficient $h(\lambda_i)=(1-s\lambda_i)^K$ determines how much the $i$-th frequency component ${\bf U}^\top_i$ contributes to the new signal~\cite{shuman2013emerging, defferrard2016convolutional, wu2019simplifying}.

\begin{figure}[t]
    \centering
    \includegraphics[width=0.8\linewidth]{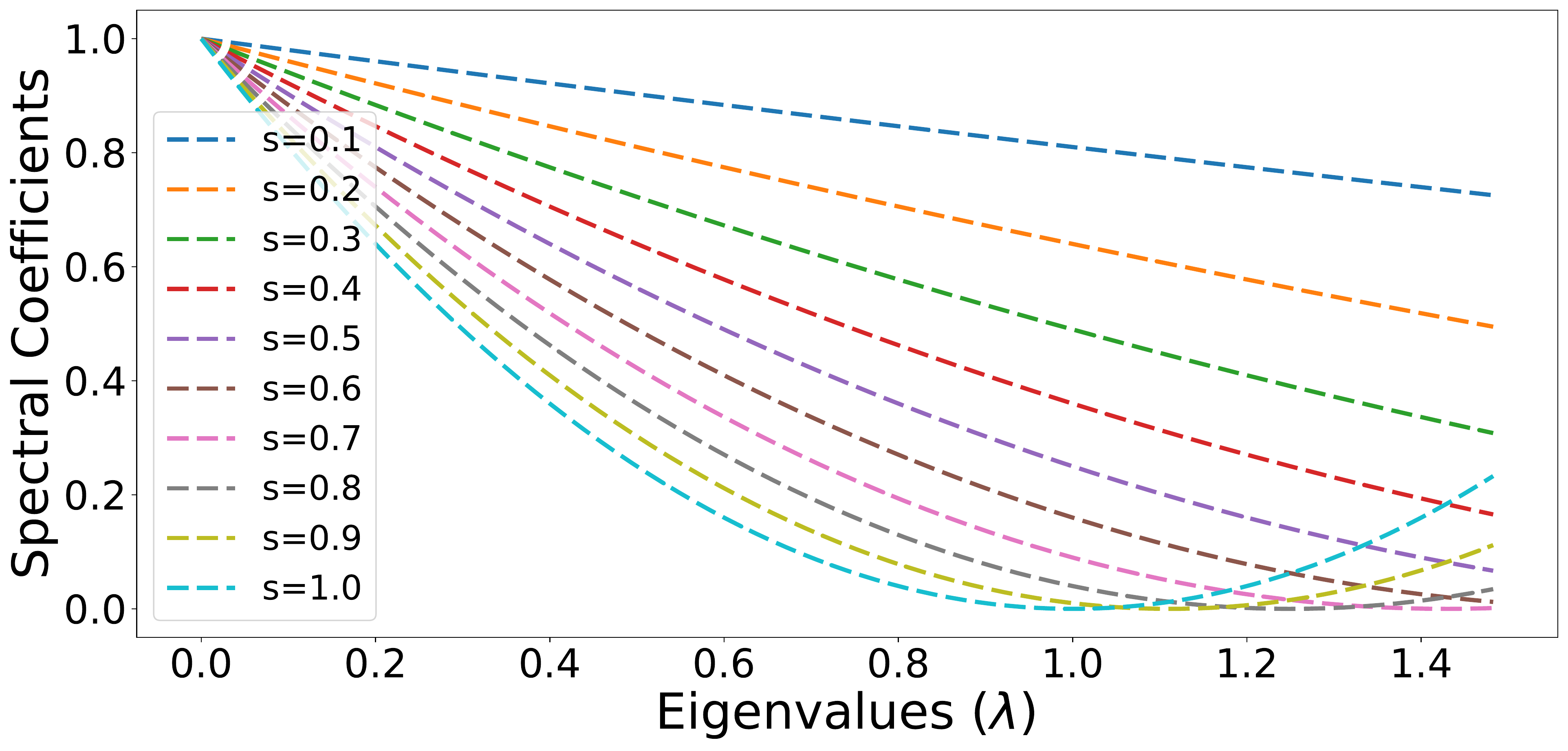}
    \vskip -0.5em
    \caption{Filter spectral coefficients on the Cora dataset.}
    \label{fig:spectral}
    \vskip -1em
\end{figure}

In Figure~\ref{fig:spectral}, we visualize $h({\lambda})$ with various $s$ ranging from $0.1$ to $1$ with an interval of $0.1$ to demonstrate how $s$ affects it. From the figure we can find that different choices of $s$ will result in different spectrum and different shape of $h(\lambda)$. By making $s$ learnable and applying different $s$ for different signal dimensions, we can obtain the same output as that in GFGN-graph as described in Section~\ref{sec:global}. It indicates that GFGN-graph is able to adjust the spectral coefficients for signal channels, thus allowing different amount information from signals of different frequencies. Meanwhile, the spectral analysis of GFGN-neighbor/pair is much more complicated since the smoothness coefficients are adaptive on different nodes. We leave this analysis as future work.

\subsection{Multi-head Gating} 
In this subsection, we compare the proposed method with GCN and analyze the number of parameters.  If we use $D^{(l)}$ and $D^{(l-1)}$ to denote the dimension of the output and input of one GCN layer, the number of parameters for one GCN layer is $O(D^{(l)}\cdot{D^{(l-1)}})$. Compared with a single GCN layer, GFGN introduces additional parameter matrix $\mathbf{W}_s\in{\mathbb{R}^{D^{(l)}\times{D}^{(l)}}}$ so the number of parameters is $O(D^{(l)}\cdot{D^{(l-1)}}+{D^{(l)}\cdot{D}^{(l)}})$. To reduce the number of parameters, we propose a multi-head gating mechanism to learn the gating scores like multi-head attention~\cite{vaswani2017attention}. Next we will use GFGN-pair to illustrate the multi-head gating mechanism.  With the mechanism, Eq.~\eqref{eq:update_pair} becomes: 

\begin{small}
\vskip -4.5ex
\begin{align}
{\bf H}^\prime_i = \bigparallel_{k=1}^{K} \sigma\left((1-\sum\limits_{v_j\in \mathcal{N}(v_i)}\frac{{\bf s}_{ij}}{d_i}){\bf H}_i {\mathbf W}^k+ \sum\limits_{v_j\in {\mathcal{N}}(v_i)}{\bf s}_{ij} \frac{{\bf H}_{j}{\mathbf W}^k}{\sqrt{{d_i}d_{j}}} \right) \nonumber
\end{align}
\end{small}where ${\bf H}^\prime_i$ is obtained by concatenating the outputs from $K$ gating heads and $\mathbf{W}^k\in{\mathbb{R}^{{D}^{(l-1)}\times\frac{D^{(l)}}{K}}}$ stands for the feature transformation matrix of the $k$-th gating head. Hence, after using multi-head gating mechanism, the complexity of the score parameter matrix in each head becomes $O(\frac{D^{(l)}}{K}\cdot\frac{D^{(l)}}{K})$ and the overall complexity becomes $O(D^{(l)}\cdot{D^{(l-1)}}+\frac{D^{(l)}}{K}\cdot{D^{(l)}})$. In practice we typically set $K=\sqrt{D^{(l)}}$ and the additional complexity becomes $O((D^{(l)})^{1.5})$. Considering the output dimension $D^{(l)}$ is often small, the additional 
number of parameters is negligible. 

\section{Experiment}
In this section, we evaluate the effectiveness of our proposed framework. In particular, we aim to answer the following questions:
(1) How does GFGN perform on various benchmark datasets compared with represenative GNN models? (2) Can GFGN combat severe random noise as expected? (3) Does the framework work as designed? 


\begin{table*}[!t]
\centering
\scriptsize
\caption{Mean accuracy (\%) over different data splits for transductive node classification. The best performance on each dataset is highlighted in bold. We report the relative improvement of our models over GCN in the last row. }
\begin{threeparttable}
\begin{tabular}{@{}lccccccccccc@{}}
\toprule
                              & \textbf{BlogCat.} & \textbf{Flickr} & \textbf{Texas} & \textbf{Wiscon.} & \textbf{Cornell} & \textbf{Actor} & \textbf{Cora}  & \textbf{Citeseer} & \textbf{Pubmed} & \textbf{ogbn-arxiv} & \textbf{\begin{tabular}[c]{@{}l@{}}Avg. \\ Rank\end{tabular}} \\ \midrule
GCN                           & 79.07                & 66.65           & 64.00          & 61.55              & 58.00            & 30.53          & \textbf{87.04} & 76.44             & 86.94   & 71.74    & 6.4                \\
GAT                           & 77.30                & 61.18           & 58.46          & 60.67              & 59.27            & 29.41          & 86.88          & 76.86             & 86.04   & 71.46        & 7.6                \\
SGC                           & 75.82                & 45.68           & 58.81          & 59.45              & 58.86            & 30.36          & 86.90          & 75.38             & 81.47   & 71.36        & 8.5                \\
ARMA                          & 93.00                & 90.78           & 74.19          & 73.45              & 67.43            & 32.44          & 85.83          & 75.91             & 86.62   &71.25        & 6.2               \\
GGNN                          & 92.95                & 89.91           & 71.73          & 71.14              & 70.27            & 34.51          & 85.28          & 74.74             & 88.93   & 71.47    & 6.2         \\  
APPNP                         & 96.06                & 91.63           & 60.76          & 63.06              & 65.81            & 34.45          & 85.11          & 75.95             & 86.65   & 71.84       & 5.9                \\
Geom-GCN*                    & N/A                  & N/A             & 67.57          & 64.12              & 60.81            & 31.63          & 85.27          & \textbf{77.99}    & \textbf{90.05}  & N/A  & 5.0                \\ \midrule
\textbf{GFGN-graph}          & 95.95                & 90.97           & \textbf{83.30}          & \textbf{85.29}             & 82.24   & 34.87          & \textbf{87.04} & 76.30             & 88.12    & 72.20      & 2.8               \\
\textbf{GFGN-neighbor}         & 95.90                & 90.85           & 82.84               & 84.29              & \textbf{82.35}            & \textbf{34.97} & 86.90          & 76.85             & 88.15   & 72.21        & 2.4                \\
\textbf{GFGN-pair}        & \textbf{96.25}       & \textbf{91.89}  & 82.65          & 83.63     & 82.27             & 34.95         & 86.74          & 76.79             & 88.12        & \textbf{72.43}     & 2.7              \\ \midrule
\textbf{\begin{tabular}[c]{@{}l@{}}Improvement\\ over GCN\end{tabular}} & 21.7\%$\uparrow$ & 37.9\%$\uparrow$ & 30.2\%$\uparrow$ & 38.6\%$\uparrow$ & 42.0\%$\uparrow$ & 14.5\%$\uparrow$ & 0 & 0.5\%$\uparrow$ & 1.4\%$\uparrow$ & 1.0\%$\uparrow$ \\
\bottomrule
\end{tabular}
\begin{tablenotes}
   \item[*] We reuse the results reported in the Geom-GCN paper~\cite{pei2020geom}. “N/A” results indicate the performances are not reported in the paper. 
\end{tablenotes}
\end{threeparttable}
\vspace{-1em}
\label{tab:node_classification}
\end{table*}
\subsection{Experimental Settings}
We conduct experiments on 11 benchmark datasets from various domains including citation, social, protein, webpage, and actor co-occurrence networks. Specifically, we use 2 social networks including BlogCatalog and Flickr~\cite{huang2017label}, 3 webpage datasets including Texas, Cornell and Wisconsin~\cite{pei2020geom}, 1 actor co-occurrence network Actor~\cite{pei2020geom}, 1 protein dataset PPI~\cite{graphsage} and 4 citation graphs including Cora, Citeseer, Pubmed~\cite{sen2008collective} and ogbn-arxiv~\cite{hu2020open-ogb}. Note that BlogCatalog, Flickr, Texas, Cornell, Wisconsin and Actor are disassortative (non-homophily) graphs where the node homophily is low, while the other 5 datasets are assoratative graphs~\cite{pei2020geom,zhu2020generalizing}. Dataset statistics can be found from Appendix A.2 in the supplementary file.


To validate our proposed method, we compare it with representative GNN models including GCN~\cite{kipf2016semi}, GAT~\cite{gat}, SGC~\cite{wu2019simplifying}, ARMA~\cite{bianchi2019graph-arma},  APPNP~\cite{klicpera2018predict-appnp} and Geom-GCN~\cite{pei2020geom}. We further include GGNN~\cite{li2015gated} which is equipped with gating systems (gated recurrent unit~\cite{cho2014learning}). We implemented our proposed method and most necessary baselines using Pytorch Geometric~\cite{fey2019fast-pyg}, a library for geometric deep learning library built upon Pytorch~\cite{paszke2017automatic-pytorch}. Detailed parameter setting and source code can be found from Appendix A.3 in the supplementary file.

\subsection{Node Classification Performance}
\label{sec:node_classification}
To answer the first question, we compare the proposed models with existing representative graph neural network models on both transductive and inductive node classification. 

\noindent\textbf{Transductive Learning.}
For the transductive learning task, we report the average accuracy on 10 random splits as well as the average rank of each algorithm in Table~\ref{tab:node_classification}. From the table, we make the following observations:

(1) The proposed three methods consistently achieve strong performance on the 10 datasets: GFGN-graph, GFGN-neighbor and GFGN-pair achieve average ranks of $2.8$, $2.4$ and $2.7$, respectively, followed by Geom-GCN with an average rank of $5.0$. Furthermore, among all the 10 datasets, the proposed methods achieve the best performance on 8 datasets. It is worth noting that our models bring in significant improvement over GCN. For example, on Flickr and Cornell datasets, our model outperforms GCN by a margin of $25.24\%$ and $24.35\%$, respectively. The superiority of the proposed methods demonstrates the importance of treating feature dimensions differently during the aggregation process of graph neural networks.

(2) Our models can bring in much more improvement on disassortative graphs than assoratative graphs. In disassorative graphs, the features between connected nodes are not as similar as those in assoratative graphs, so the methods (GCN, GAT and SGC) based on local aggregation will aggregate noisy information from neighbors, thus resulting in node embeddings of bad quality. On the contrary, as described in Section~\ref{sec:framework}, our models can learn different smoothing scores for different feature dimensions, which enables them to adaptively balance the information from the node itself and its neighbors. Specifically, for two connected nodes, if most of their features are dissimilar, our methods can learn to capture such information and assign small smoothness scores for those feature dimensions. For the detailed analysis of the learned smoothness scores, we will further investigate them in Section~\ref{sec:score_analysis}. Although our model does not improve GCN on assortative graphs as significantly as on disassortative graphs, the results show that the proposed aggregation scheme does not hurt the performance and is comparable with existing representative graph neural networks.

(3) Different datasets could need different-level smoothing strategies. For example, GFGN-graph significantly outperforms the other two GFGN variants on Wisconsin dataset, indicating that graph-level smoothing is good enough for improving the performance on this dataset while the other two might not be well trained since they need to learn more coefficients. In contrast, GFGN-pair works the best on Flickr dataset, suggesting that the finest level of smoothing (pair-level) is important on Flickr. 

\noindent\textbf{Inductive Learning.} We then evaluate the effectiveness of GFGN on PPI dataset for inductive learning. We compare GFGN with previous SOTA methods on PPI and   report their performance in Table~\ref{tab:ppi}. All three GFGN variants achieve competitive performance and GFGN-neighbor outperforms other methods by a large margin. It suggests that enabling different contributions for different feature dimensions is also necessary for the task of inductive learning.

\begin{table}[th]
\small
\centering
\caption{Micro-averaged F1 scores on PPI.}
\begin{tabular}{@{}ll@{}}
\toprule
Methods     & F1 score   \\ \midrule
GraphSAGE~\cite{graphsage}   & 61.2  \\
VR-GCN~\cite{chen2017stochastic}      & 97.8  \\
GaAN~\cite{zhang2018gaan}        & 98.71 \\
GAT~\cite{gat}         & 97.3  \\
GeniePath~\cite{liu2019geniepath}   & 98.5  \\
GraphSAINT~\cite{zeng2019graphsaint}  & 98.1 \\
\midrule
GFGN-graph & 99.27$\pm$ 0.02\\
GFGN-neighbor & \textbf{99.32$\pm$ 0.01} \\
GFGN-pair & 98.40$\pm$ 0.04\\
\bottomrule
\end{tabular}
\label{tab:ppi}
\vspace{-1em}
\end{table}

\begin{figure*}[tb]%
     \centering
     \subfloat[Cora]{{\includegraphics[width=0.3\linewidth]{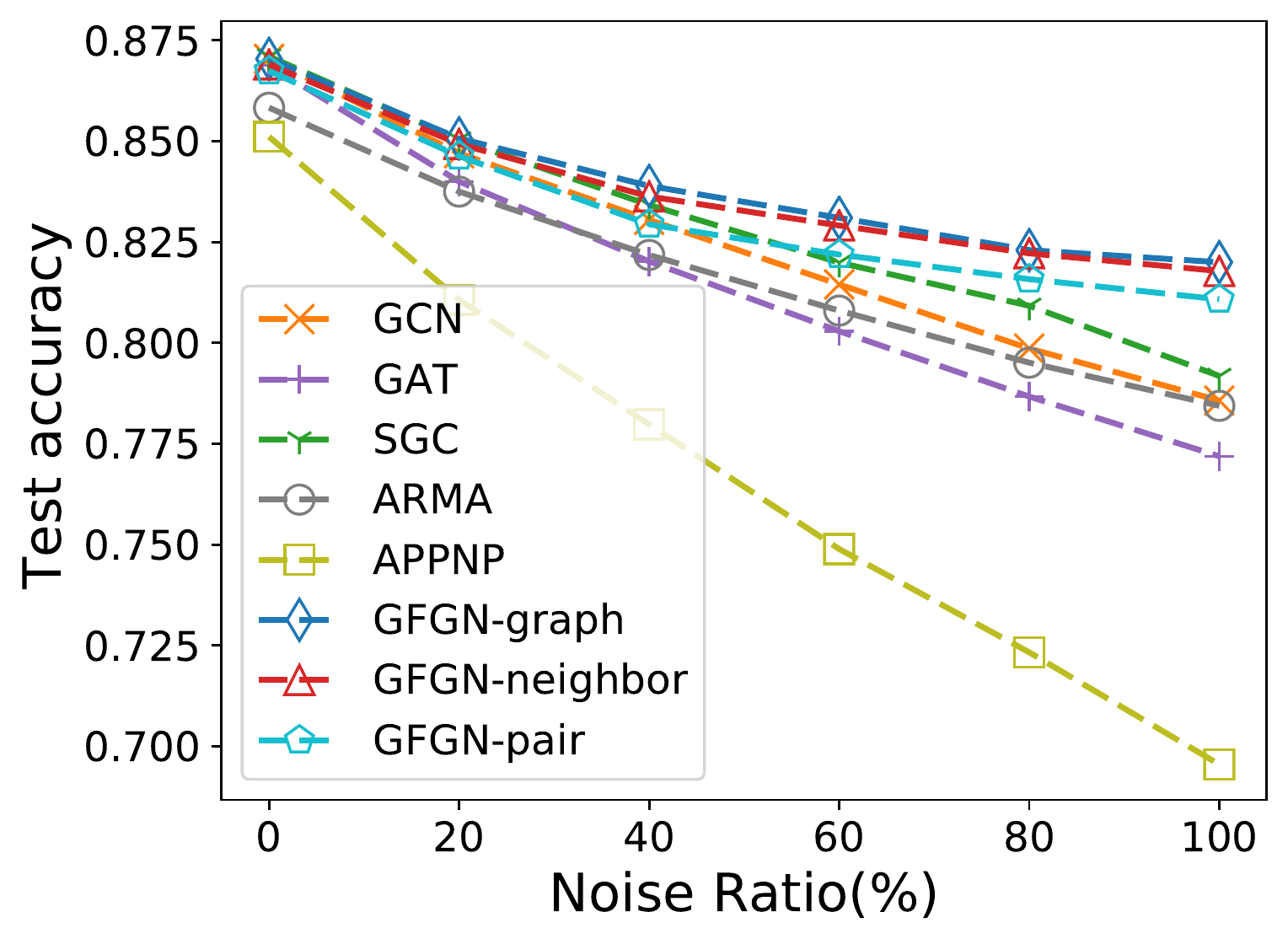} }}%
     \subfloat[Citeseer]{{\includegraphics[width=0.3\linewidth]{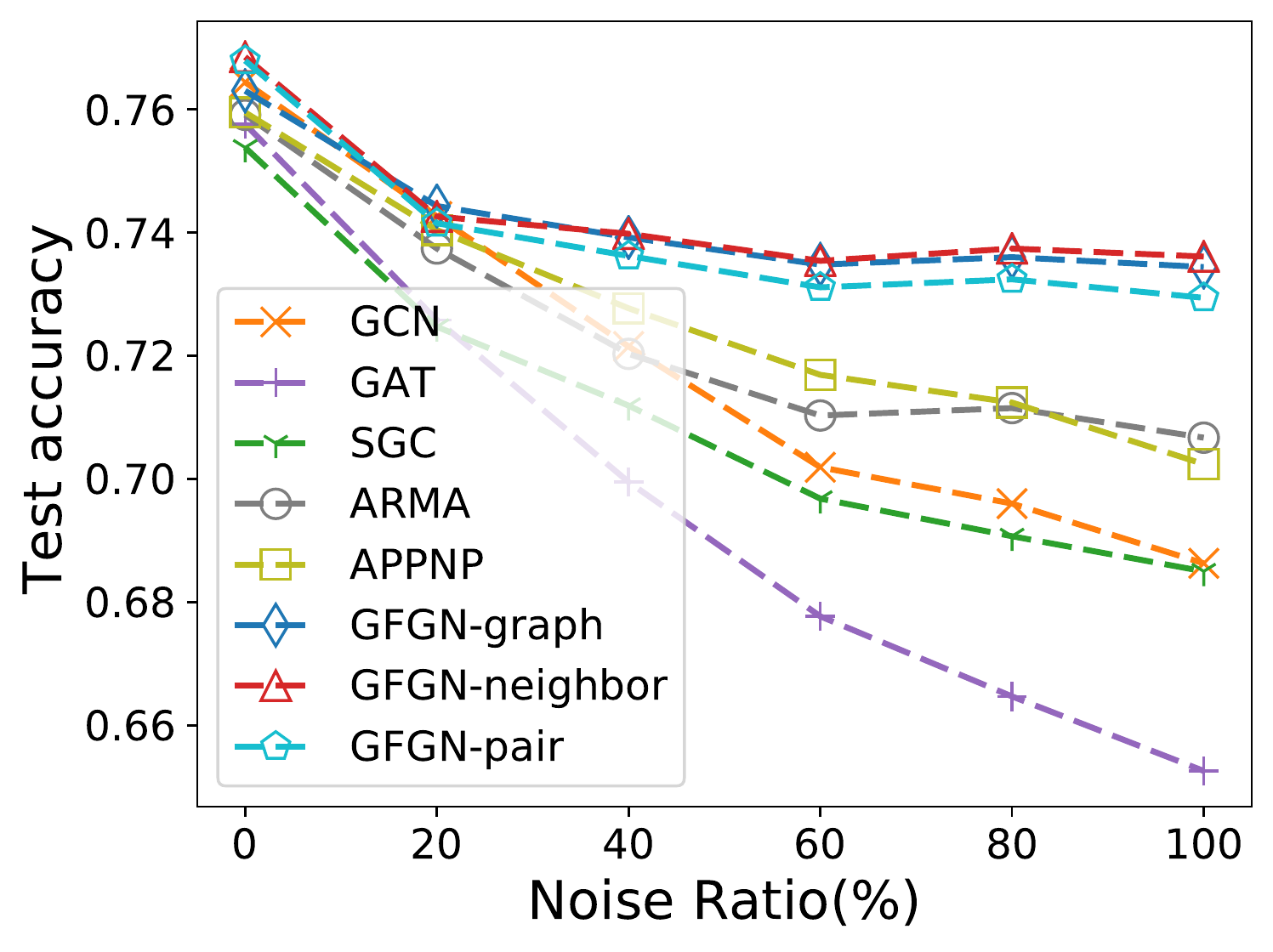} }}%
    \subfloat[Pubmed]{{\includegraphics[width=0.3\linewidth]{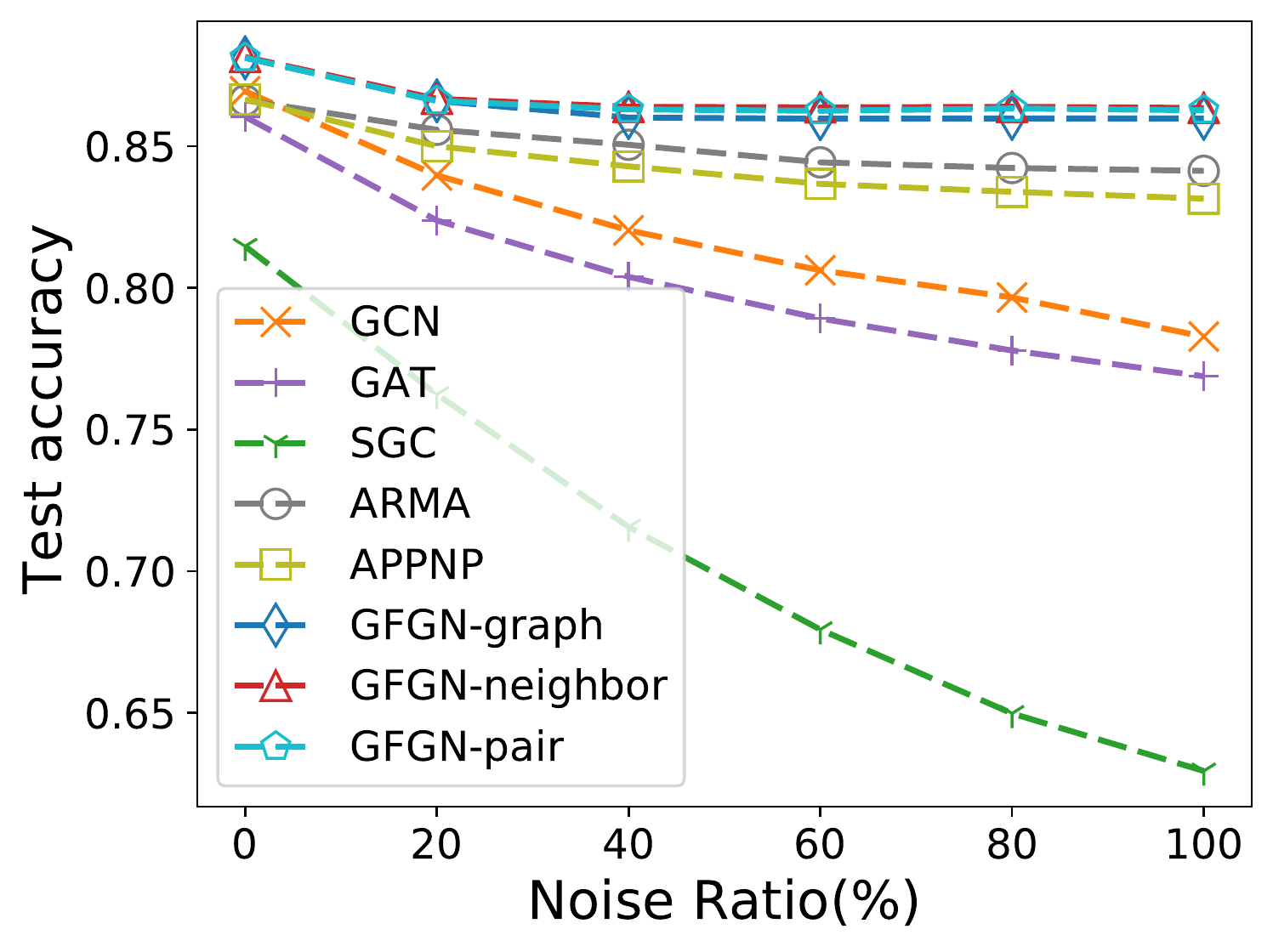} }}%
    \qquad
    \vskip -0.4em
    \caption{Node classification accuracy under different ratio of random noise.}
    \label{fig:robustness}
\vskip -0.5em
\end{figure*}

\subsection{Robustness Under Random Noise}
\label{sec:random_noise}
When the graph structure is injected with random noise, a lot of new connections will be created. 
Considering the huge amount of the possible node pairs in the graph, most of the random connections will connect nodes with totally different features and some connect nodes with partially dissimilar features. 
The connections that connect totally/partially dissimilar nodes will inevitably bring in noisy information when traditional GNN models aggregate information from neighbors, leading to performance deterioration. In contrast, our proposed GFGN can learn to balance the information from neighbors and the node itself as well as balancing the information from different feature dimensions, thus it has the potential to boost model robustness under random noise. Specifically, for the connections that connect partially dissimilar nodes, GFGN is expected to learn low smoothing scores for those dissimilar feature dimensions while learning high smoothing scores for those similar dimensions.

In this subsection, we aim to evaluate how our model behaves under different degrees of random noises. Specifically, we conduct the experiments on Cora, Citeseer and Pubmed datasets. Specifically, we inject random noise into the graph structure by randomly adding a certain ratio of edges. We vary the ratio from 0\% to 100\% with a step size of 20\% 
and report the node classification accuracy in Figure~\ref{fig:robustness}. The figure shows that the proposed three graph filters can consistently outperform all other baselines and our methods are more robust under random noise. As we can see from the figure, GFGN consistently outperforms other baselines under different degrees of random noise on all the three datasets. Specifically, under $100\%$ random noise, GFGN improves GCN by a margin of $3.5\%$, $4.8\%$ and $8.1\%$ on Cora, Citeseer, Pubmed, respectively. It demonstrates that the proposed GFGN is more robust under random noise than traditional graph neural network models.




\begin{figure*}[ht]%
\vskip -0.2em
    \centering
    \subfloat[BlogCatalog]
        {{\includegraphics[width=0.25\linewidth]{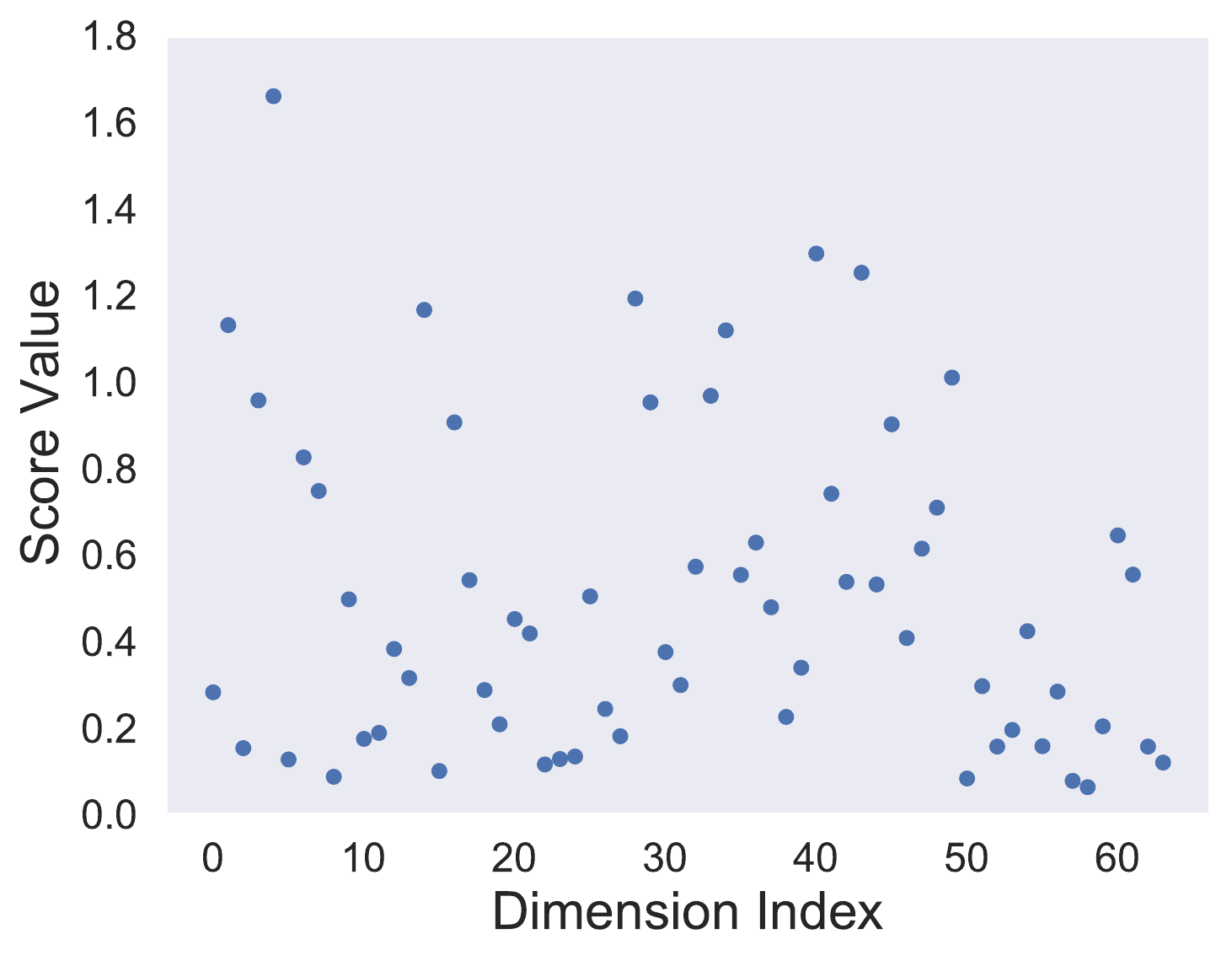}}}%
    \subfloat[Cora]{{\includegraphics[width=0.25\linewidth]{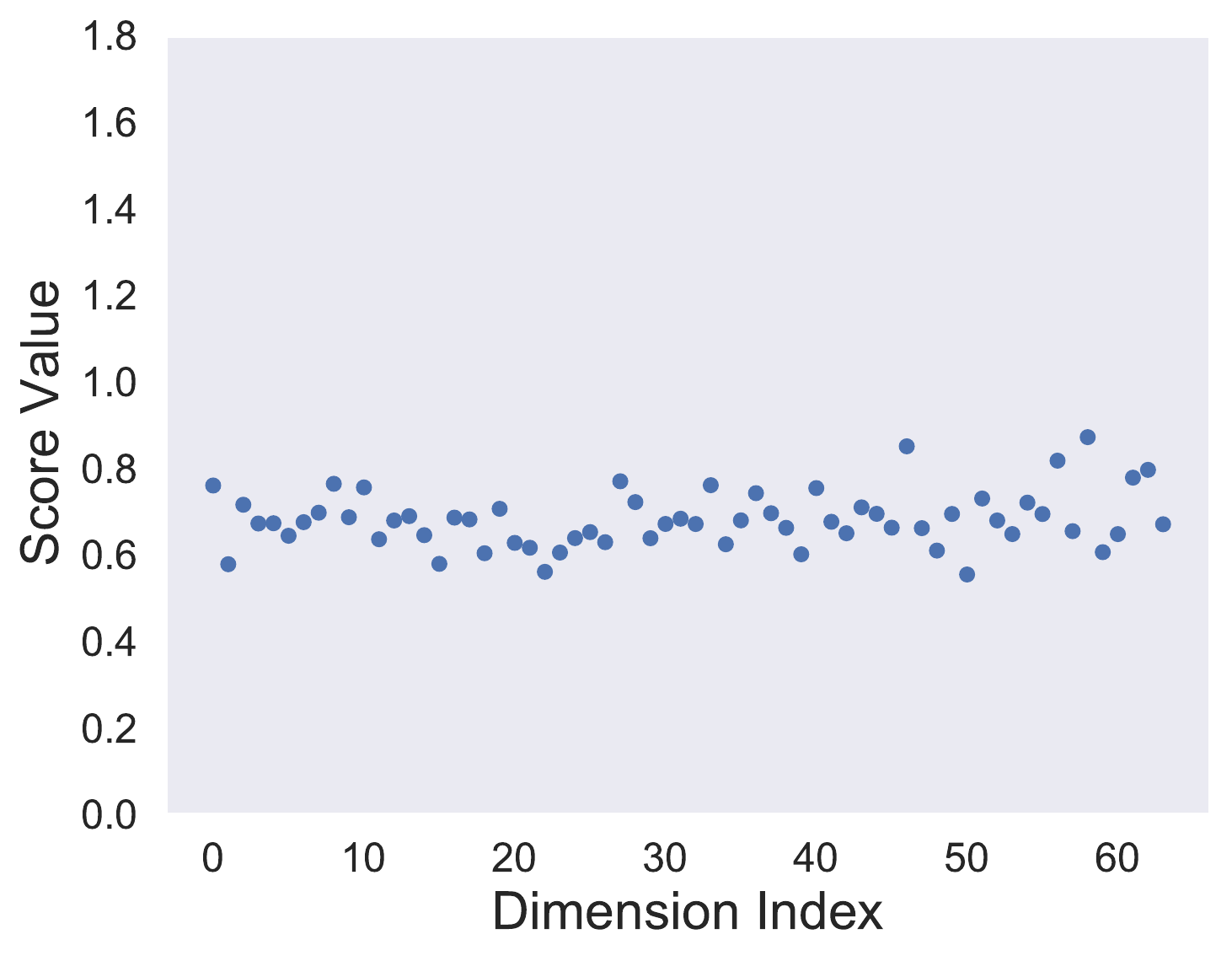} }}%
    \subfloat[BlogCatalog]{{\includegraphics[width=0.25\linewidth]{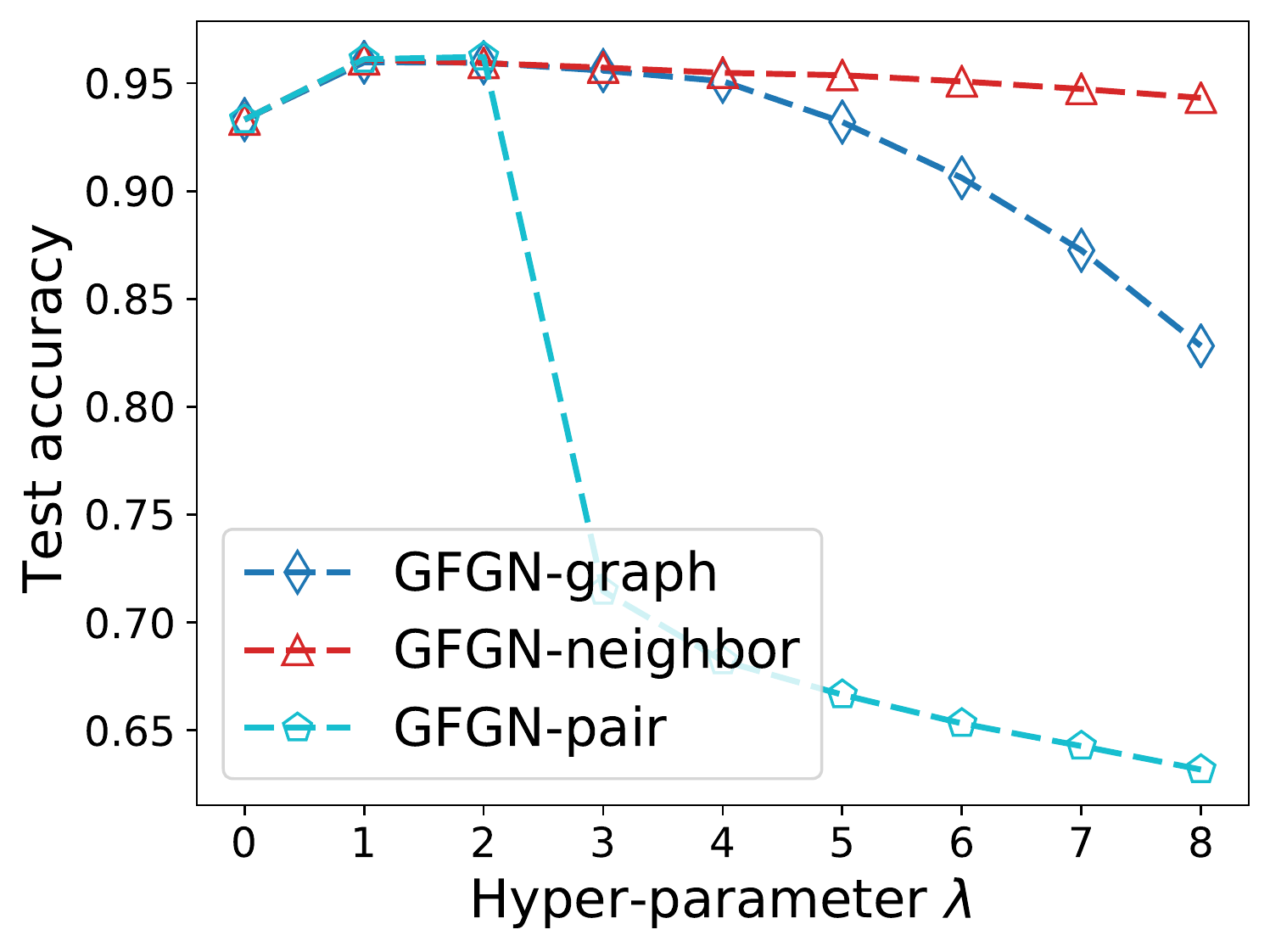} }}%
    \subfloat[Cora]{{\includegraphics[width=0.25\linewidth]{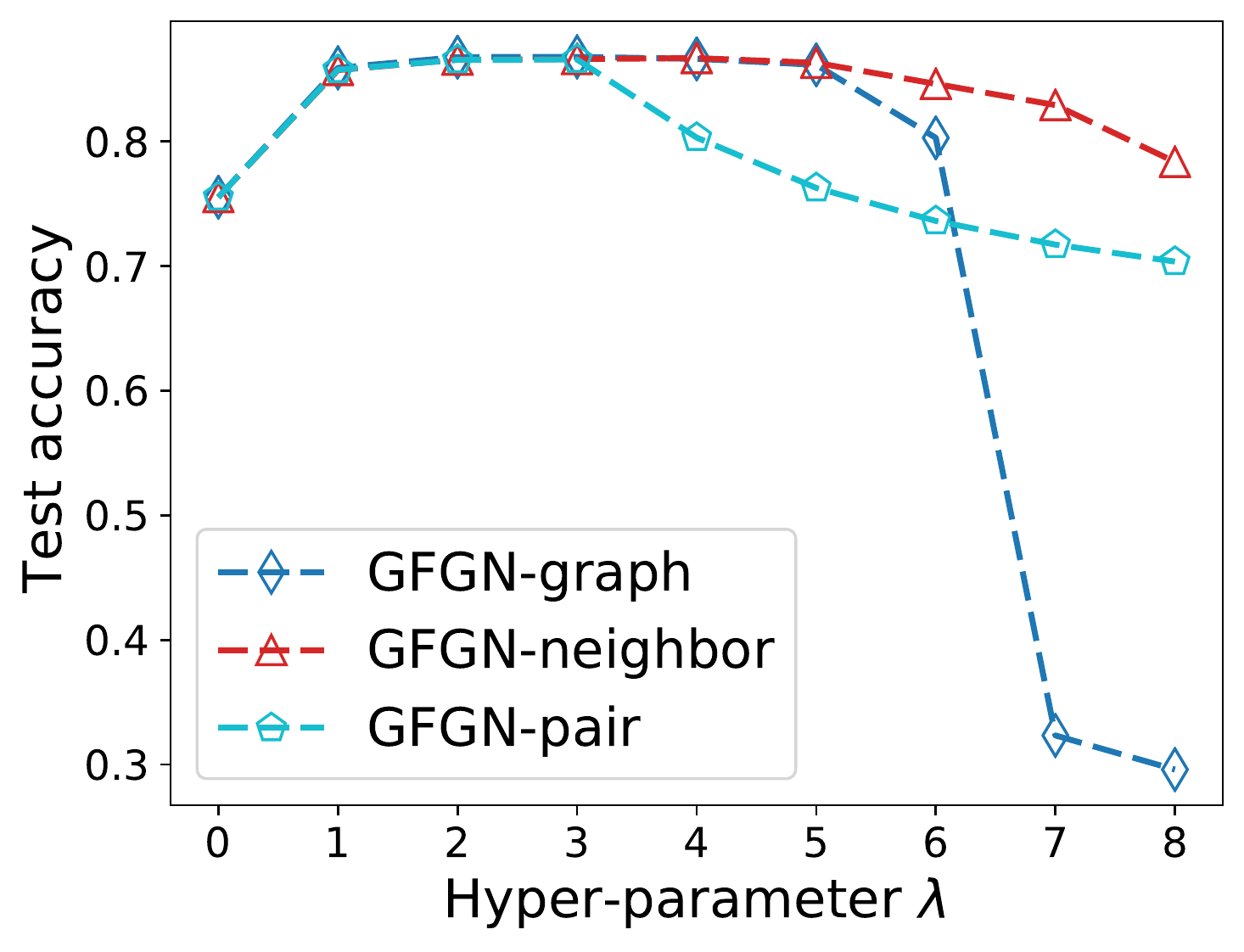} }}%
    \qquad
    \caption{\textbf{(a)(b)}The values of learned smoothing scores in the first layer of GFGN-graph. \textbf{(c)(d)} Parameter analysis.}%
    \label{fig:scores}
    \vskip -0.5em
\end{figure*}


\subsection{Deeper Understanding Towards GFGN}
\label{sec:score_analysis}
In this subsection, we perform analysis on the smoothing scores and hyper-parameters to answer the third question.

\textbf{The Learned Smoothing Scores.} Due to the page limit, we focus on studying GFGN-graph and provide the results for GFGN-neighbor and GFGN-pair in Appendix A.4 of the supplementary file while we note that similar trends can be found for them. Specifically, we choose the smoothing scores from the first layer of GFGN-graph and plot the values with respect to different feature dimensions in Figure~\ref{fig:scores} for BlogCatalog and Cora while similar patterns are exhibited in other datasets. From the results, we can make the following observations:

\noindent(1) The learned smoothing scores in all datasets vary in some range, indicating that our model can learn different scores for different feature dimensions. 
Furthermore, the variance of the scores in disassorative graphs is much larger than these in assortative graphs. More specifically, the learned scores in BlogCatalog vary in the range of $[0.06, 1.66]$, while in Cora the range is only $[0.55, 0.87]$. It could be the reason that almost all the feature dimensions satisfy the homophliy assumption in assortative graphs but only a portion of dimensions satisfy it in disassortative graphs. Despite the low homophily of disassorative graphs, GFGN manages to exploit the homogeneous information in some feature dimensions and adaptively learns high smoothing scores for them while assigning low scores to other dimensions. 

\noindent(2) The smoothing scores learned for disassortative graphs tend to be smaller than the scores in assorative graphs. On BlogCatalog, we find that 64.1\% of the score values are smaller than 0.55, respectively, while 100\% of the values on Cora are larger than 0.55. This phenomenon indicates that feature information in those disassortative graphs is more important than structural information such that we should give low smoothing scores in most of the feature dimensions and aggregate less information from neighbors.

\textbf{Parameter Analysis.}
\label{sec:params}
We  explore how the value of  hyper-parameter $\lambda$ impacts the performance of the three variants of GFGN. We vary its value and report the results on BlogCatalog and Cora in Figure~\ref{fig:scores}. 
Note that when $\lambda=0$, GFGN reduces to multilayer perceptron (MLP). We make the following observations:

(1) When choosing an appropriate value of $\lambda$, GFGN consistently outperforms MLP which does not incorporate the graph structural information. However, a large value of $\lambda$ can hurt the performance. This could be the reason that a large $\lambda$ allows too much freedom in learning the smoothing scores and makes the training process unstable.  

(2) Among the three variants, GFGN-neighbor tend to be more stable than the other two. Specifically, when the value of $\lambda$ increases, GFGN-graph is very sensitive and the performance drops rapidly; GFGN-pair also suffers from the same issue. GFGN-graph assigns the same smoothing score vector for all nodes, which is not flexible for the individuals; GFGN-pair requires to learn a score for each node pair and it becomes harder to learn good scores when the graph is dense (e.g. BlogCatalog). Compared with them, GFGN-neighbor can learn adaptive scores for each node and the number of score vectors is relatively small. 

\section{Related Work}
Graph neural networks (GNNs) have been attracting increasing attention in recent years, due to their great capability for learning representations on graphs. In general, there are majorly two types of GNN models, i.e. spectral-based and spatial-based. The spectral-based GNNs are built upon graph convolution operations defined in the spectral domain~\cite{shuman2013emerging}. Together with linear transform and non-linearity, the graph convolution operation can be utilized to form graph convolution layers, which are typically stacked to form GNN models~\cite{bruna2013spectral, henaff2015deep,defferrard2016convolutional}. On the other hand, spatial-based GNNs have been proposed way before  the spectral-based ones, which dates back to the time when deep learning was not yet popular. However, its development has stagnated since then until the emergence of GCN~\cite{kipf2016semi}, which is a simplified spectral-based model and also can be treated as a spatial-based model. More spatial-based GNNs have since been developed~\cite{gat,hamilton2017inductive, gilmer2017neural}. The spatial-based graph convolution can be understood as propagating features across the graph, which is simple yet effective  in various applications~\cite{ying2018graph,yan2018spatial,marcheggiani2018exploiting,zitnik2018modeling}. 
Furthermore, there are some advanced topics in GNNs such as deep graph neural networks~\cite{liu2020towards-oversmooth}, self-supervised graph neural networks~\cite{you2020does,jin2020selfsupervised,You2020GraphCL}, robust graph neural networks~\cite{jin2020graph,pagnn,jin2020adversarial} and explainability on graphs~\cite{liu2021dig}. For a thorough review, we please refer the reader to a recent book~\cite{ma2020deep}. 

Most of existing works utilize a unified way to propagate features for all feature dimensions, which may not be optimal since different dimensions of features may represent different accept of nodes and thus have different smoothness. In this paper, we try to address this issue by introducing gating system to control smoothness for each dimension from perspectives of graph-level, neighborhood-level, and pair-level, respectively. Note that, though gating systems have been adopted for GNNs in a few previous works~\cite{li2015gated,ruiz2019gated,bresson2017residual}, they are used as tools for controlling information flow through layers. In contrast, we use gating systems as tools to control the smoothness for different feature dimensions. 



\section{Conclusion}
Graph neural networks are powerful tools for graph representation learning. However, they follow a neighborhood aggregation scheme which simply treats each feature dimension equally. Motivated by the property of spectral embedding and social dimension theory, we propose a novel general aggregation scheme where each feature dimension is considered differently during aggregation by learning different smoothing scores to different dimensions through gating units. Based on the general framework, we introduce three graph filters to capture different levels of feature smoothing, i.e., graph-level, neighborhood-level and pair-level. Extensive experiments have demonstrated the effectiveness of the proposed methods on various benchmark datasets. We further show that the proposed GFGN is more robust to severe random noise compared with other GNN models. In the future, we aim to explore the spectral properties of the proposed neighborhood-level and pair-level smoothing.

\balance
\bibliography{sample}
\bibliographystyle{icml2021}

\appendix
\section{Appendix}
\subsection{Proof for Motivation Example 2}

{\bf Motivation Example 2: Spectral Embedding.} By selecting top eigenvectors of the graph Laplacian matrix as node representations, the method of spectral embedding embeds the graph into a low-dimensional space where each dimension corresponds to an eigenvector~\cite{von2007tutorial,ng2002spectral}. Specifically, given the adjacency matrix $\mathbf{A}$ with $\mathbf{D}$ as its degree matrix, its $k$-th eigenvector $\mathbf{f}_k\in\mathbb{R}^{N\times{1}}$ can be obtained through $({\bf D} - \mathbf{A}){\mathbf f}_k = \lambda_k \mathbf{f}_k$ with $\lambda_k$ being the $k$-th eigenvalue. In fact, $\mathbf{f}_k$ denotes the $k$-th dimension of node representations (or the vector with the $k$-th feature of all nodes).
The smoothness of feature embedding $\mathbf{f}_k$ can be measured by $k$-th eigenvalue of $\mathbf{D}-\mathbf{A}$:
\begin{align}
    \sum_{i,j}\|\mathbf{f}_k(v_i) - \mathbf{f}_k(v_j)\|^2 &= \mathbf{f}^{\top}_k (\mathbf{D}-\mathbf{A})\mathbf{f}_k \nonumber \\
    &= \lambda_k \|\mathbf{f}_k\|^2 \nonumber  \\ 
    &= \lambda_k
    \label{eq:spectral_smoothness}
\end{align}
if $\|\mathbf{f}_k\|^2=1$.
Since $\lambda_k$ is different for each dimension, Eq.~\eqref{eq:spectral_smoothness} suggests that spectral embedding treats each feature dimension differently
and $\lambda_k$ controls the smoothness of the feature dimension $\mathbf{f}_k$ over the graph. 

\begin{table}[h]
\scriptsize
\caption{Dataset statistics. The first 6 datasets are disassortative graphs while the last 5 datasets are assortative graphs.}
\begin{tabular}{@{}lccccc@{}}
\toprule
Datasets    & \#Nodes & \#Edges & \#Features & \#Classes & Task \\
\midrule 
BlogCatalog & 5,196    & 171,743  & 8,189       & 6      & Transductive    \\
Flickr      & 7,575    & 239,738  & 12,047      & 9      & Transductive    \\
Actor       & 7,600    & 33,544   & 931        & 5       & Transductive   \\
Cornell     & 183     & 295     & 1,703       & 5        & Transductive  \\
Texas       & 183     & 309     & 1,703       & 5        & Transductive  \\
Wisconsin   & 251     & 499     & 1,703       & 5        & Transductive  \\ \midrule
Cora        & 2,708    & 5,429    & 1,433       & 7      & Transductive    \\
Citeseer    & 3,327    & 4,732    & 3,703       & 6      & Transductive    \\
Pubmed      & 19,717   & 44,338   & 500        & 3       & Transductive  \\  
ogbn-arxiv & 169,343 & 1,166,243 & 128  & 40 &  Transductive \\
PPI & 56,944 & 818,716 & 50 & 121 & Inductive \\
\bottomrule
\end{tabular}
\vspace{-1em}
\label{tab:dataset}
\end{table}

\subsection{Code and Dataset}
Our code can be found in the attached supplementary files. Dataset statistics are summarized in Table~\ref{tab:dataset}.
Following the experimental setting of Geom-GCN~\cite{pei2020geom}, for all datasets except ogbn-arxiv and PPI, we use 10 random splits, choosing 48\% of the nodes for training, 32\% validation and 20\% for test.
For ogbn-arxiv and PPI, we follow the public splits provided by the literature. For each random split, the experiments are repeated 10 times with different random seeds. For ogbn-arxiv dataset, we use the public splits provided in~\cite{hu2020open-ogb} where the split ratio for training/validation/test is 54\%/18\%/28\%. For inductive experiments on PPI dataset (which contains 24 graphs),  we follow the setting of previous work~\cite{gat}, to use 20 graphs for training, 2 graphs for validation, and the rest for test. 



\subsection{Parameter Setting}
\label{sec:setup}
For a fair comparison, all methods are evaluated on the same data splits. For all methods, we tune hyperparameters individually on each dataset based on the validation set. 
For ogbn-arxiv, we adopt 4-layer GFGN with 16 heads, 16 hidden units in each head, and learning rate $0.003$. For PPI, we apply 6-layer GFGN layers with 32 heads, 64 hidden units and learning rate 5$e$-4. For other datasets, we set the number of GFGN layers to 2, the number of heads to 8, the number of hidden units to 8 and learning rate 0.005 or 0.05. Further, we search dropout rate from $\{0.5, 0.8\}$, $\lambda$ from $\{1, 2\}$ and $L_2$ normalization from $\{$5$e$-4, 5$e$-5$\}$.

\subsection{Visualization of Smoothing Scores}
\begin{figure*}[!htb]%
\vskip -0.2em
    \centering
    \subfloat[BlogCatalog]
        {{\includegraphics[width=0.25\linewidth]{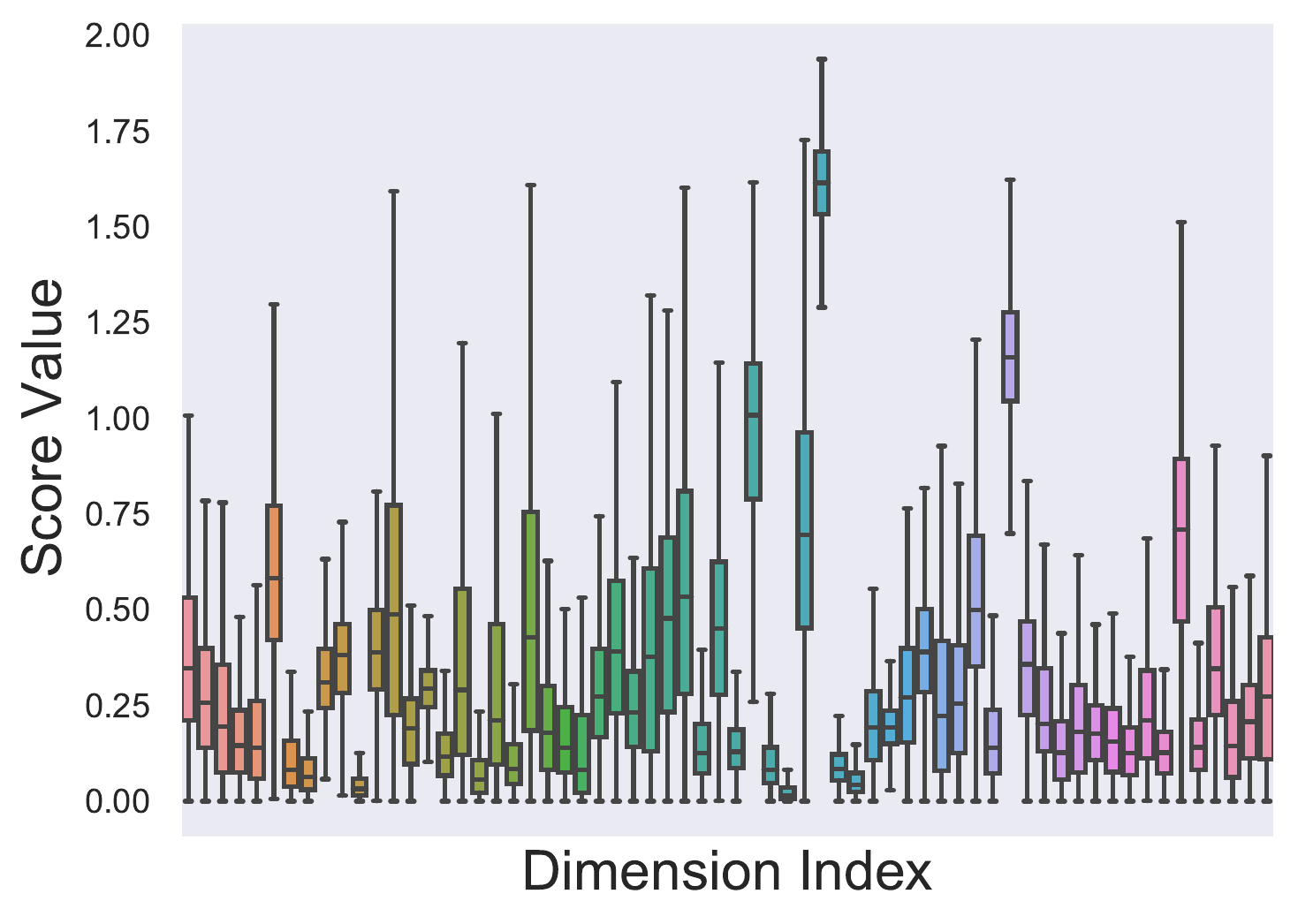}}}%
    \subfloat[Flickr]{{\includegraphics[width=0.25\linewidth]{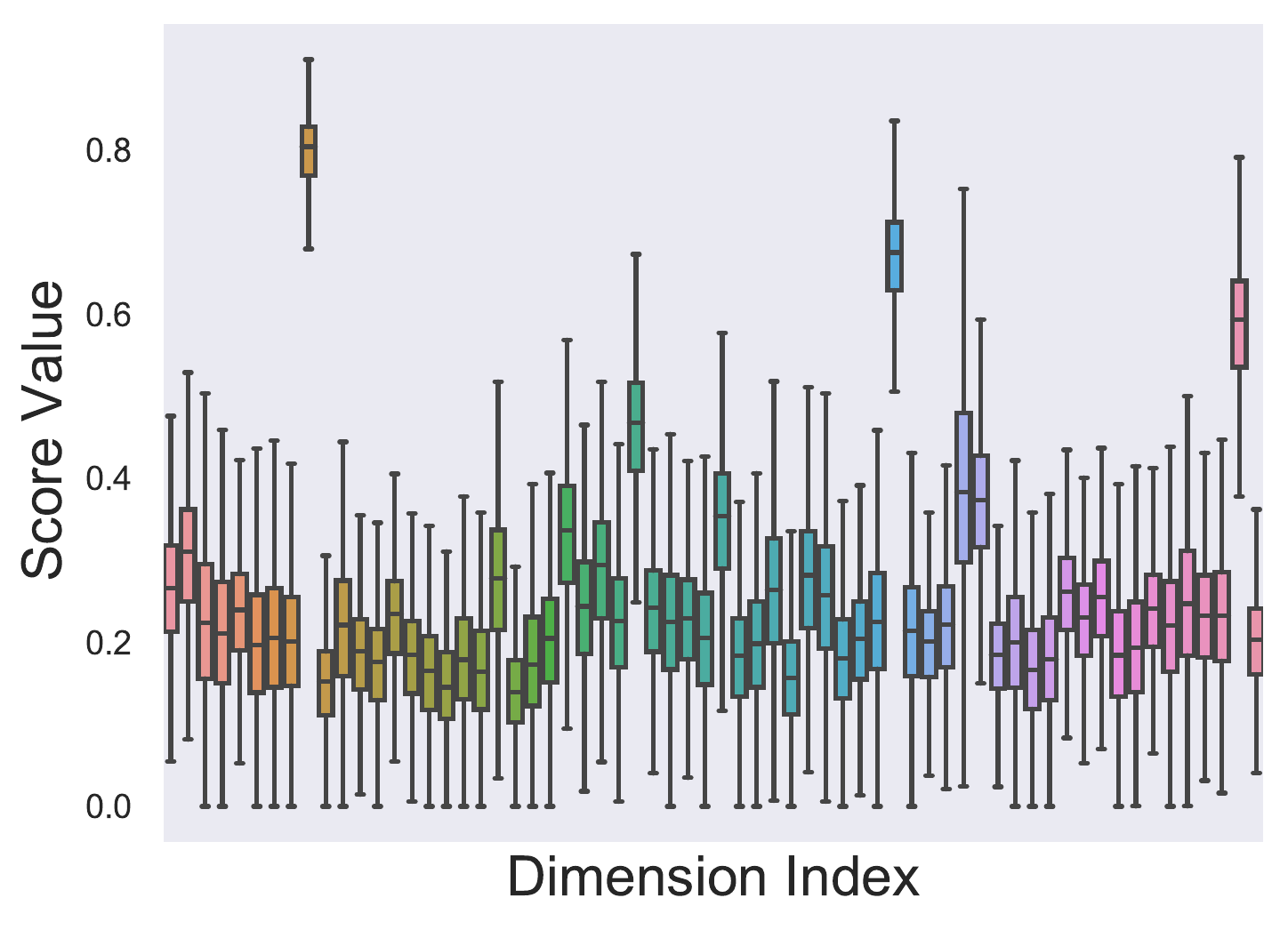} }}%
    \subfloat[Texas]{{\includegraphics[width=0.25\linewidth]{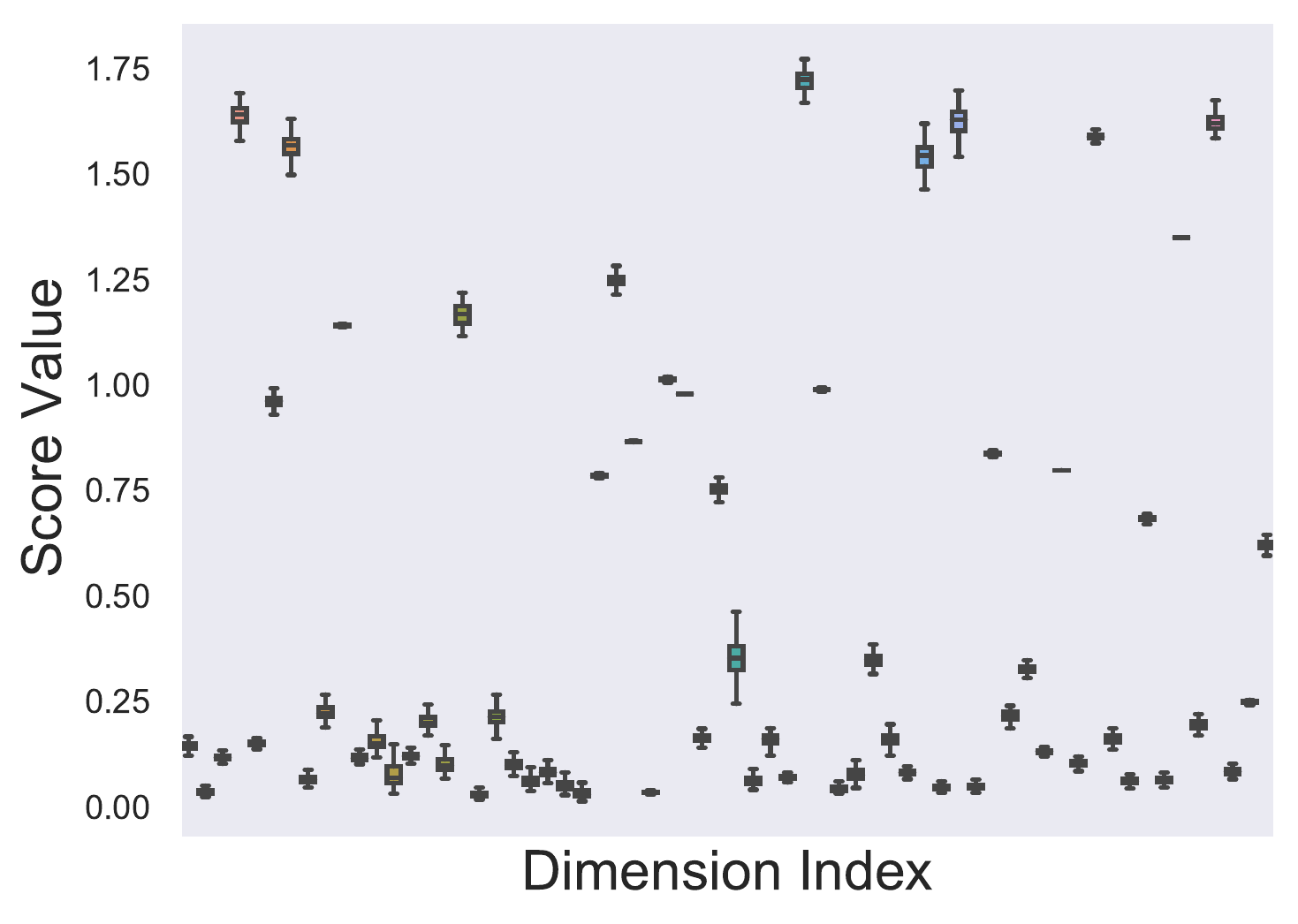} }}%
    \subfloat[Wisconsin]
        {{\includegraphics[width=0.25\linewidth]{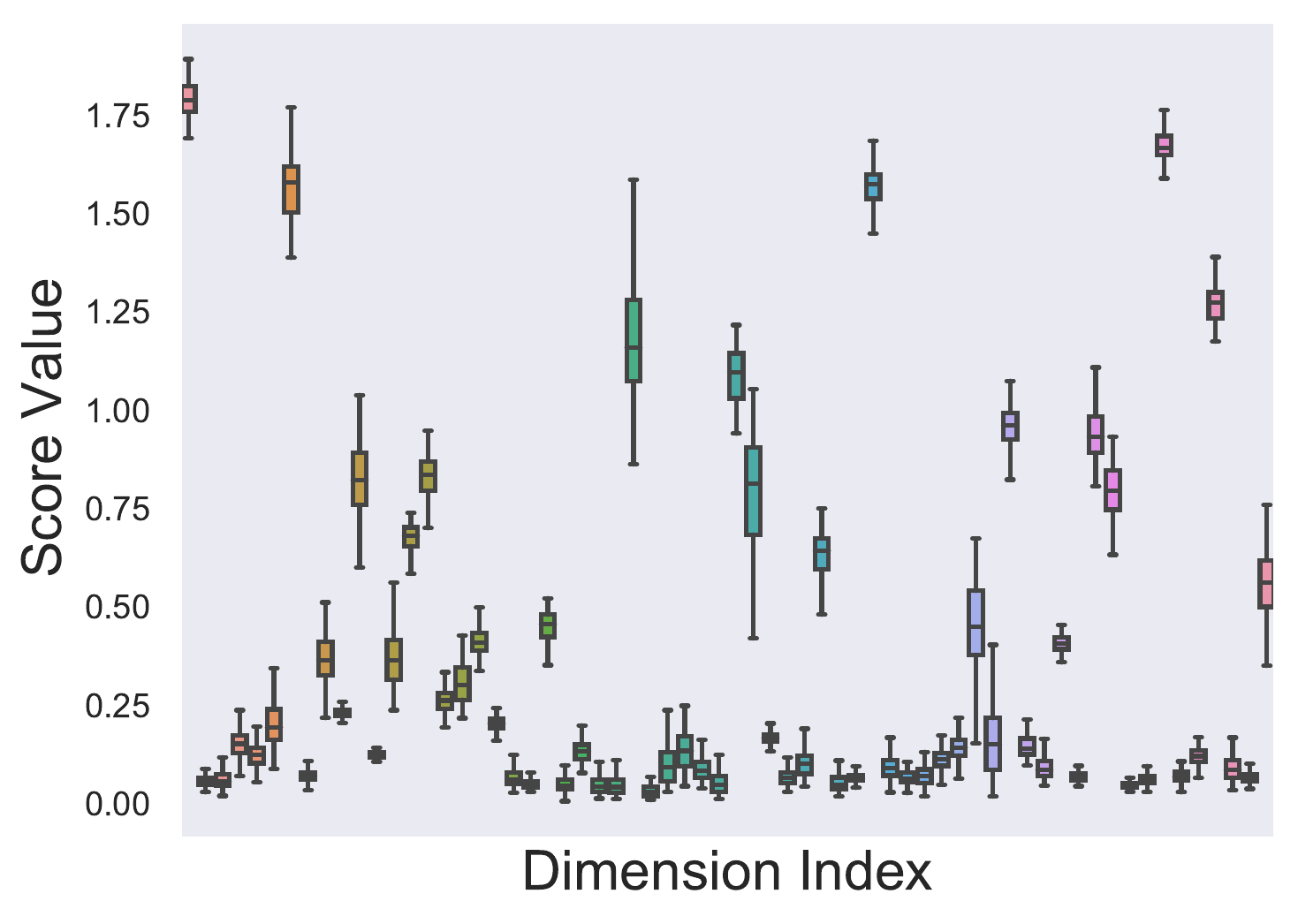}}}%
    \qquad
    \subfloat[Cornell]{{\includegraphics[width=0.25\linewidth]{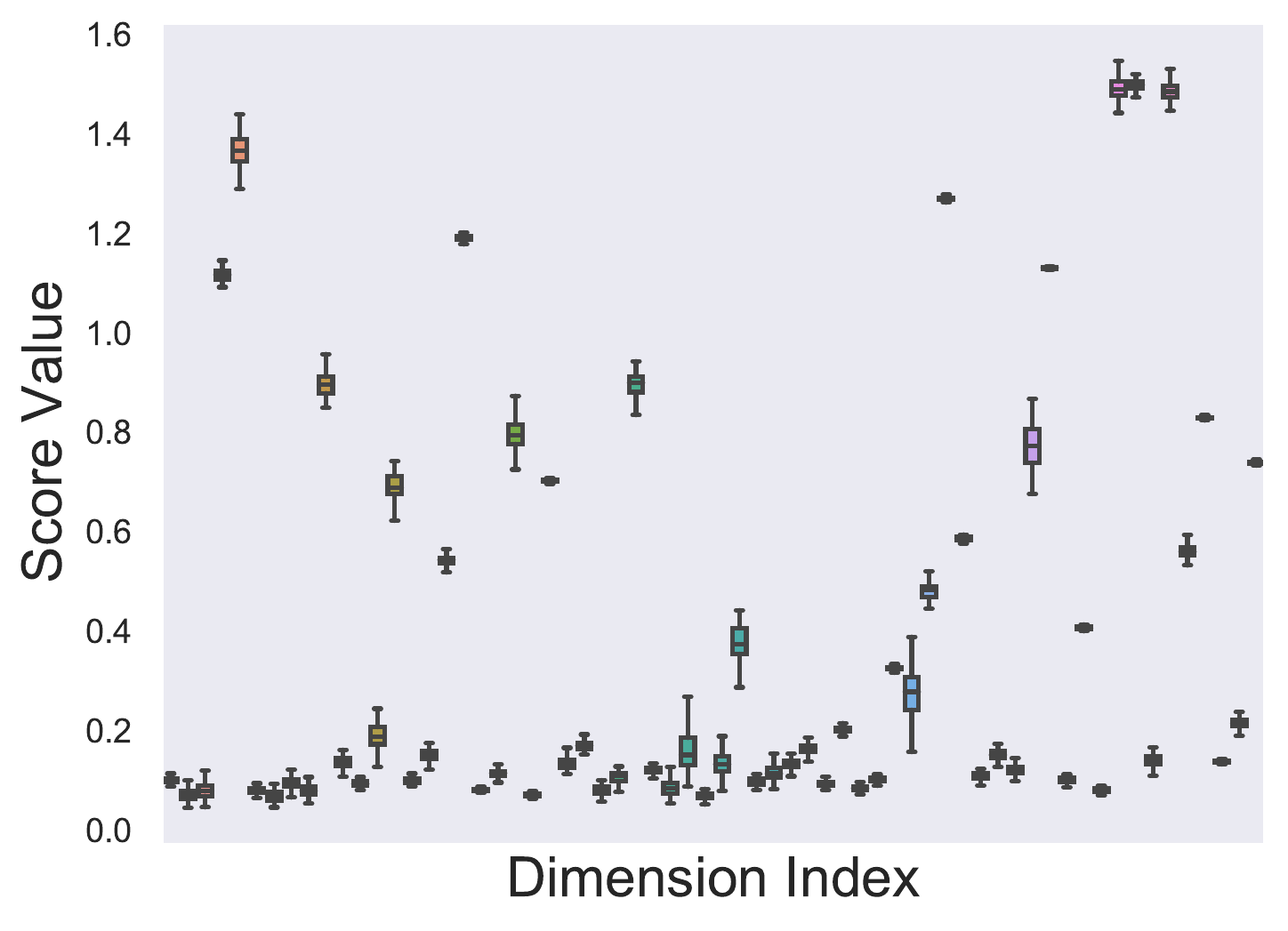}}}%
    \subfloat[Cora]
        {{\includegraphics[width=0.25\linewidth]{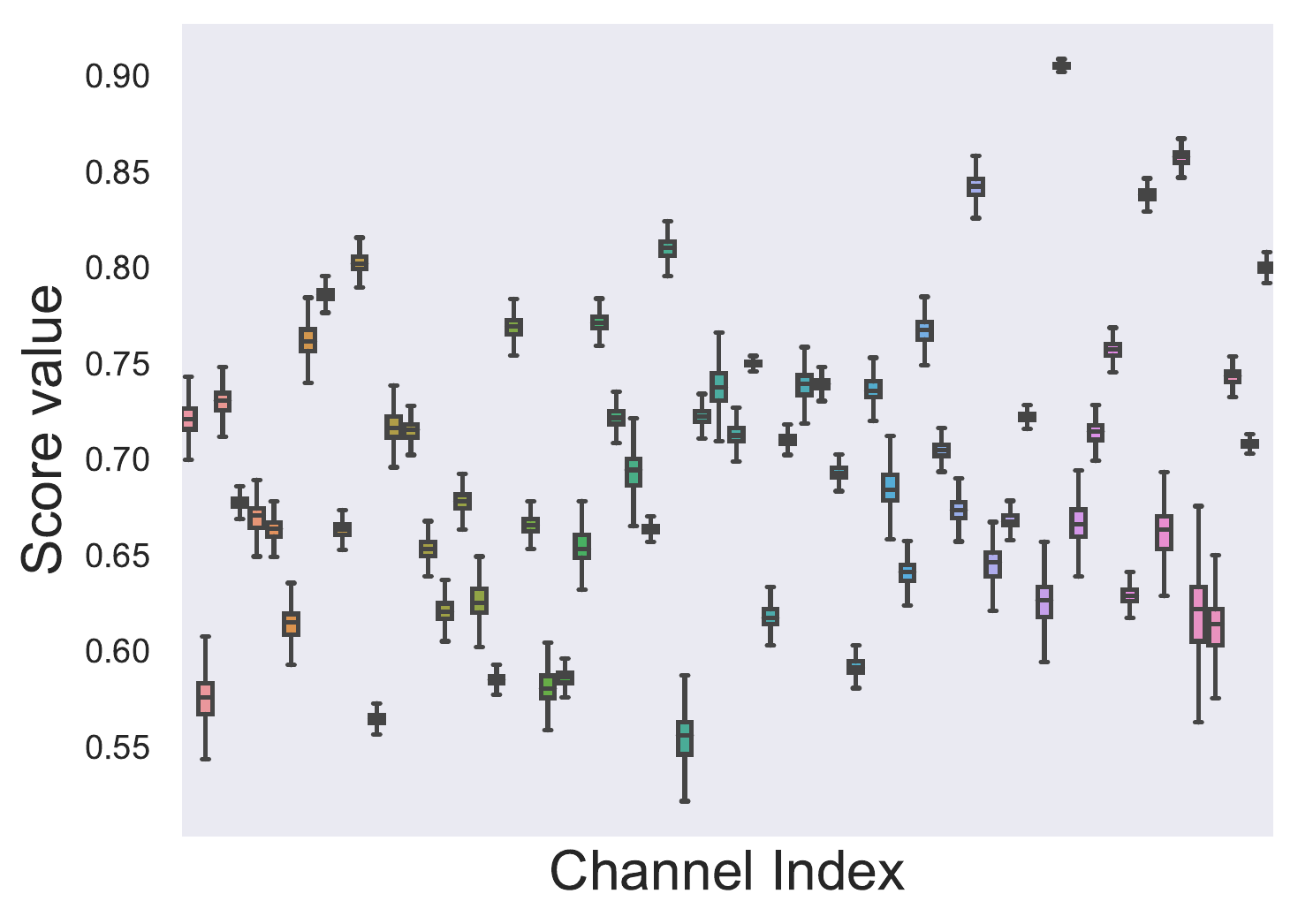}}}%
    \subfloat[Citeseer]{{\includegraphics[width=0.25\linewidth]{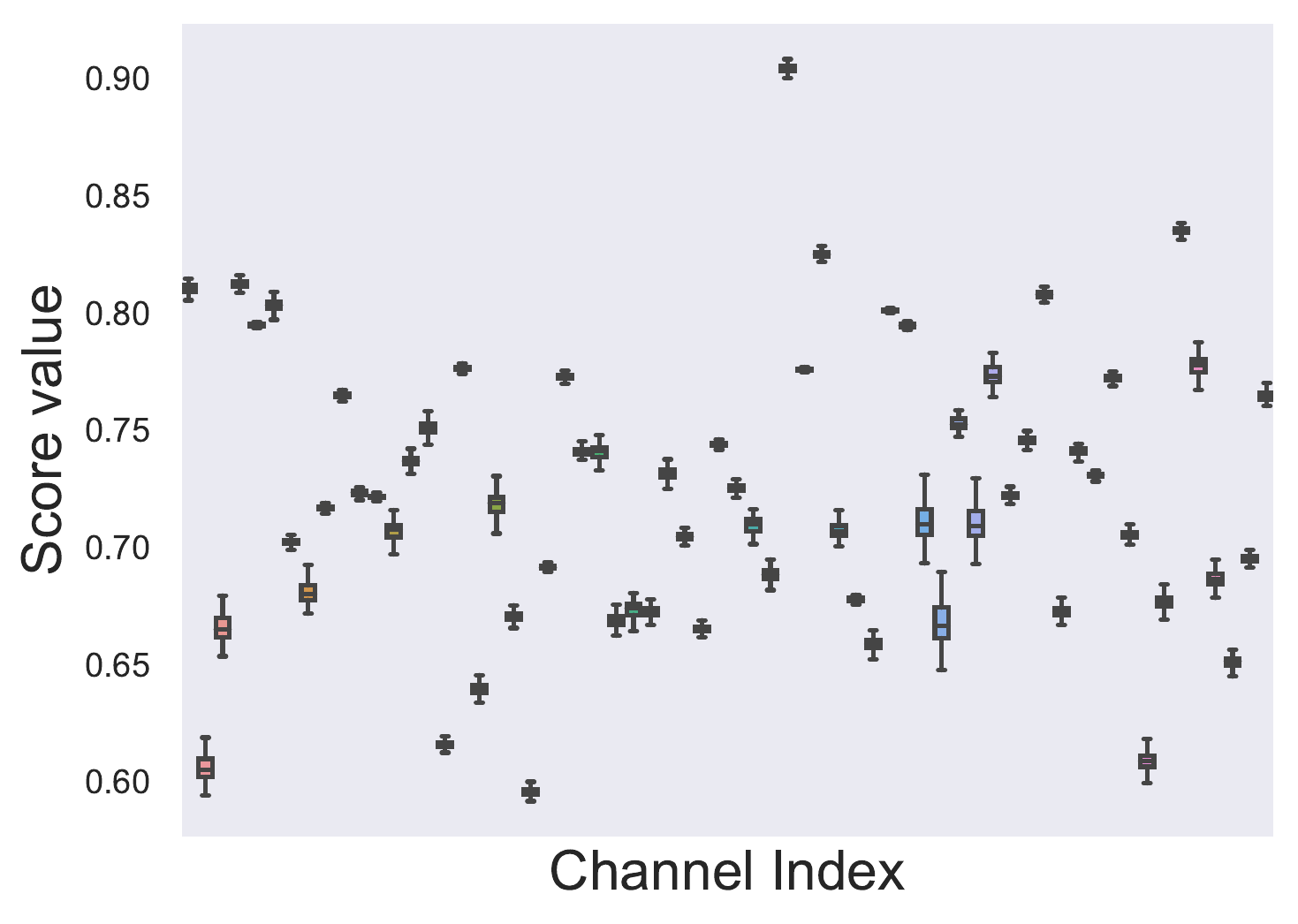}}}%
    \subfloat[Pubmed]{{\includegraphics[width=0.25\linewidth]{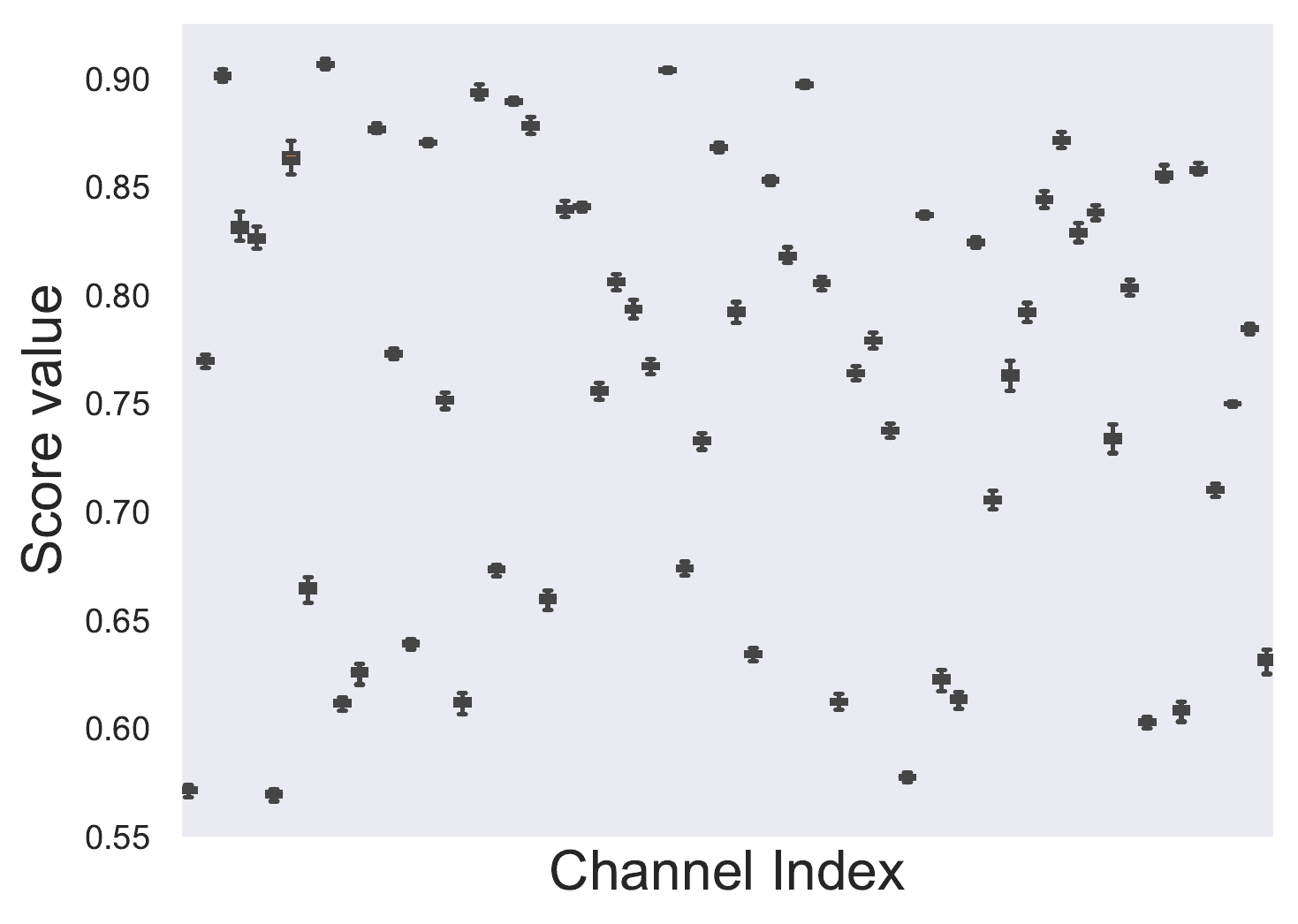} }}%
    \qquad
    \vskip -0.6em
    \caption{The values of learned smoothing scores in the first layer of GFGN-neighbor.}%
    \label{fig:scores_neighbor}
    \vskip -1em
\end{figure*}

\begin{figure*}[!htb]%

    \centering
    \subfloat[BlogCatalog]
        {{\includegraphics[width=0.25\linewidth]{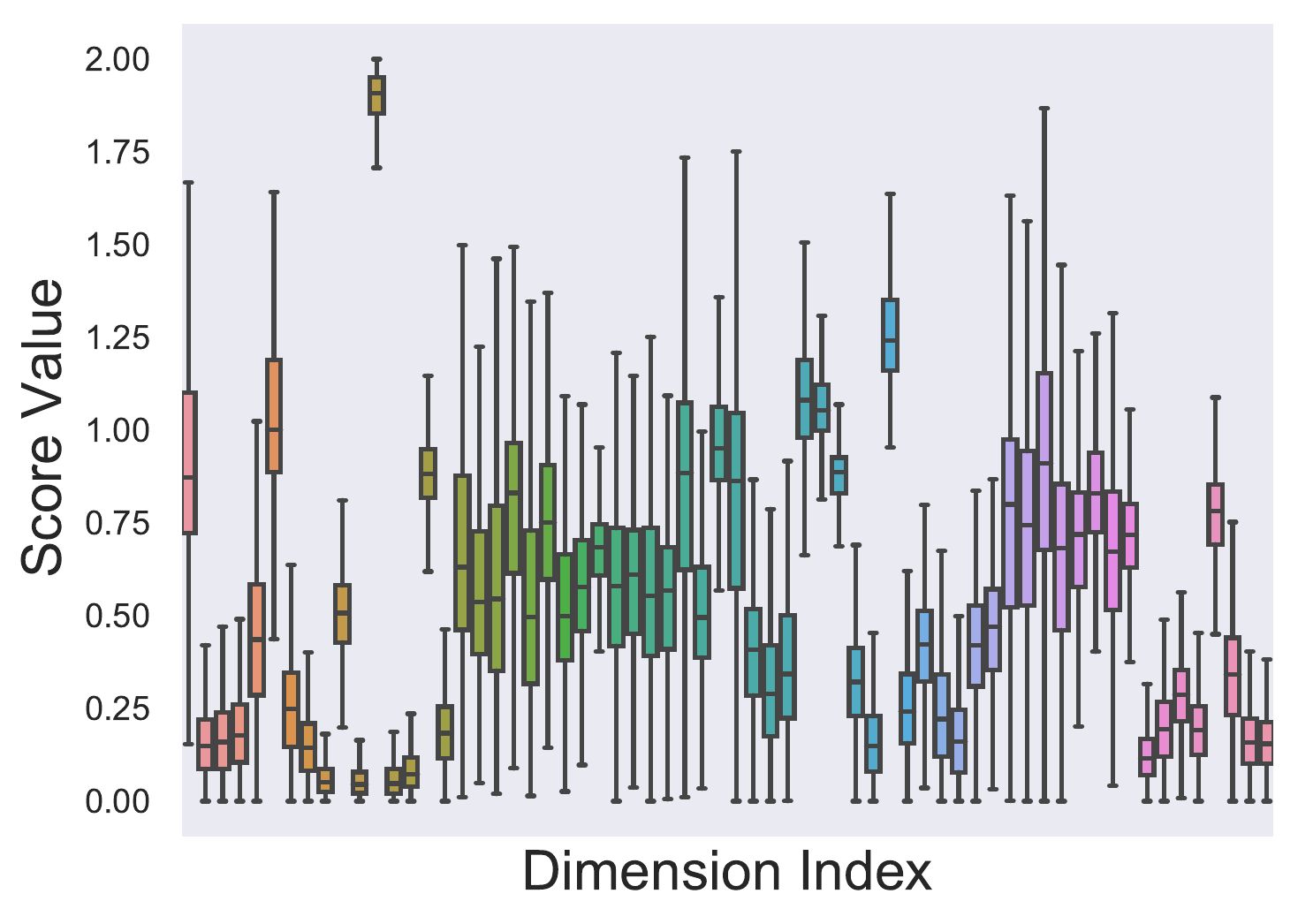}}}%
    \subfloat[Flickr]{{\includegraphics[width=0.25\linewidth]{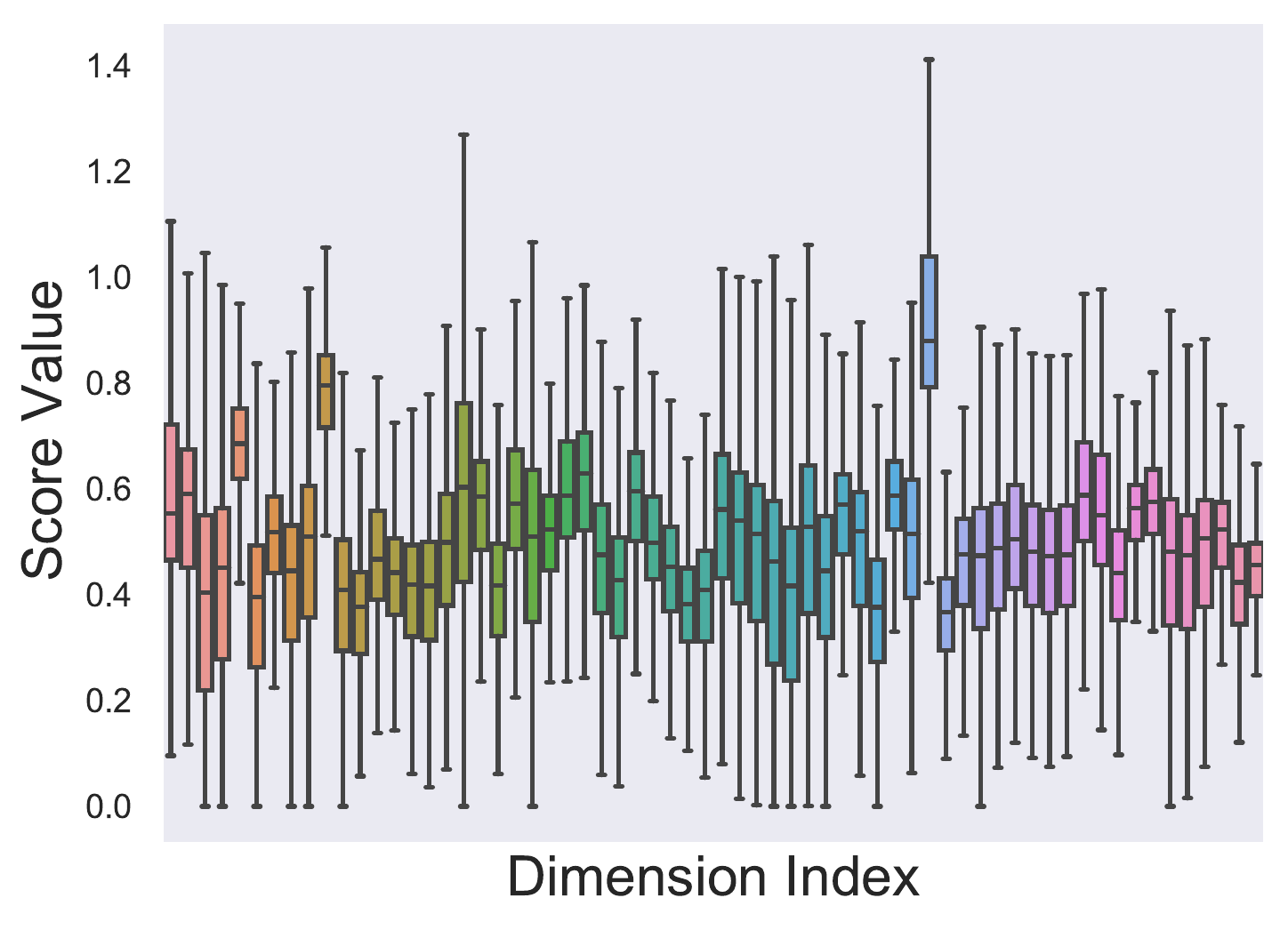} }}%
    \subfloat[Texas]{{\includegraphics[width=0.25\linewidth]{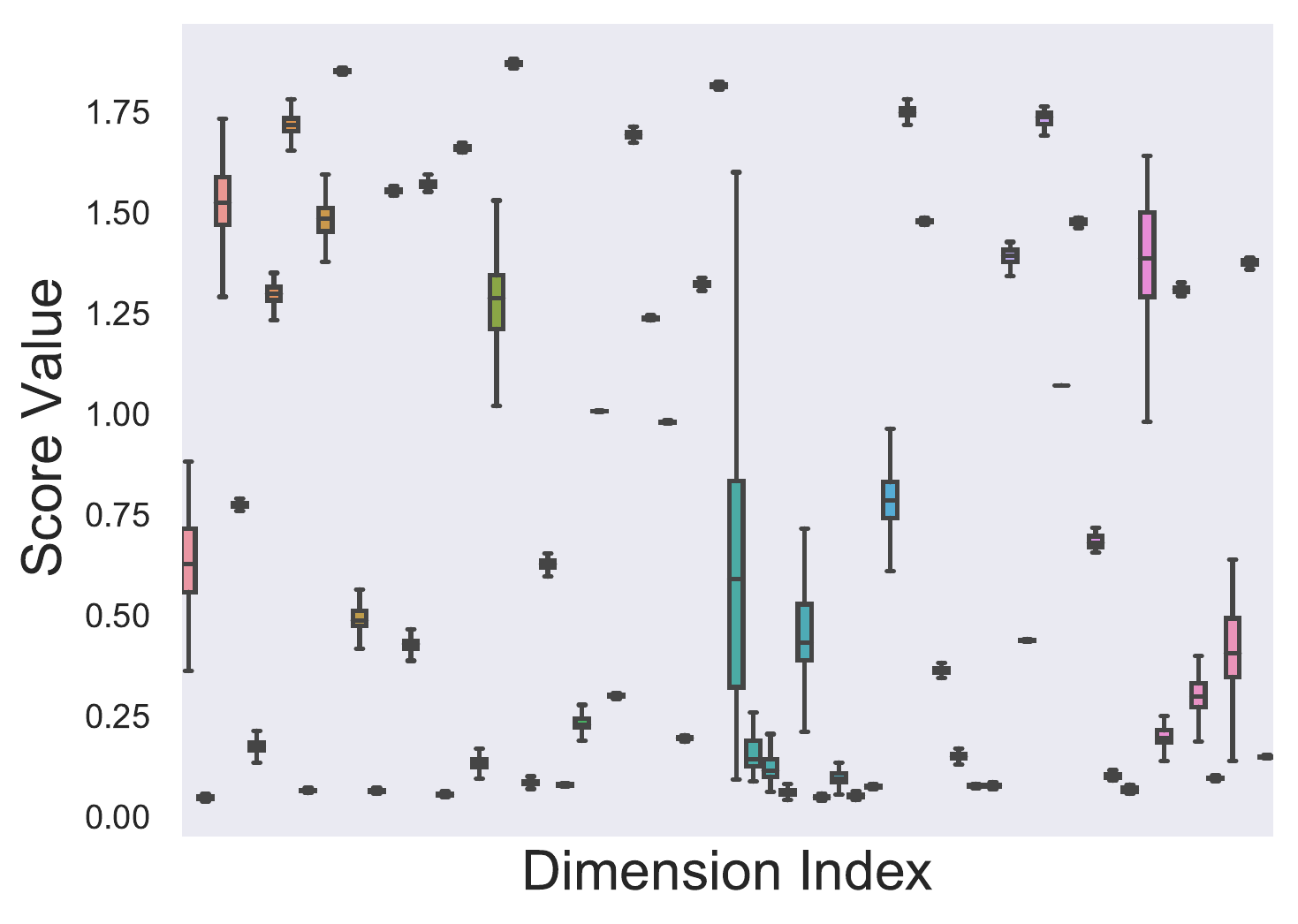} }}%
    \subfloat[Cora]
        {{\includegraphics[width=0.25\linewidth]{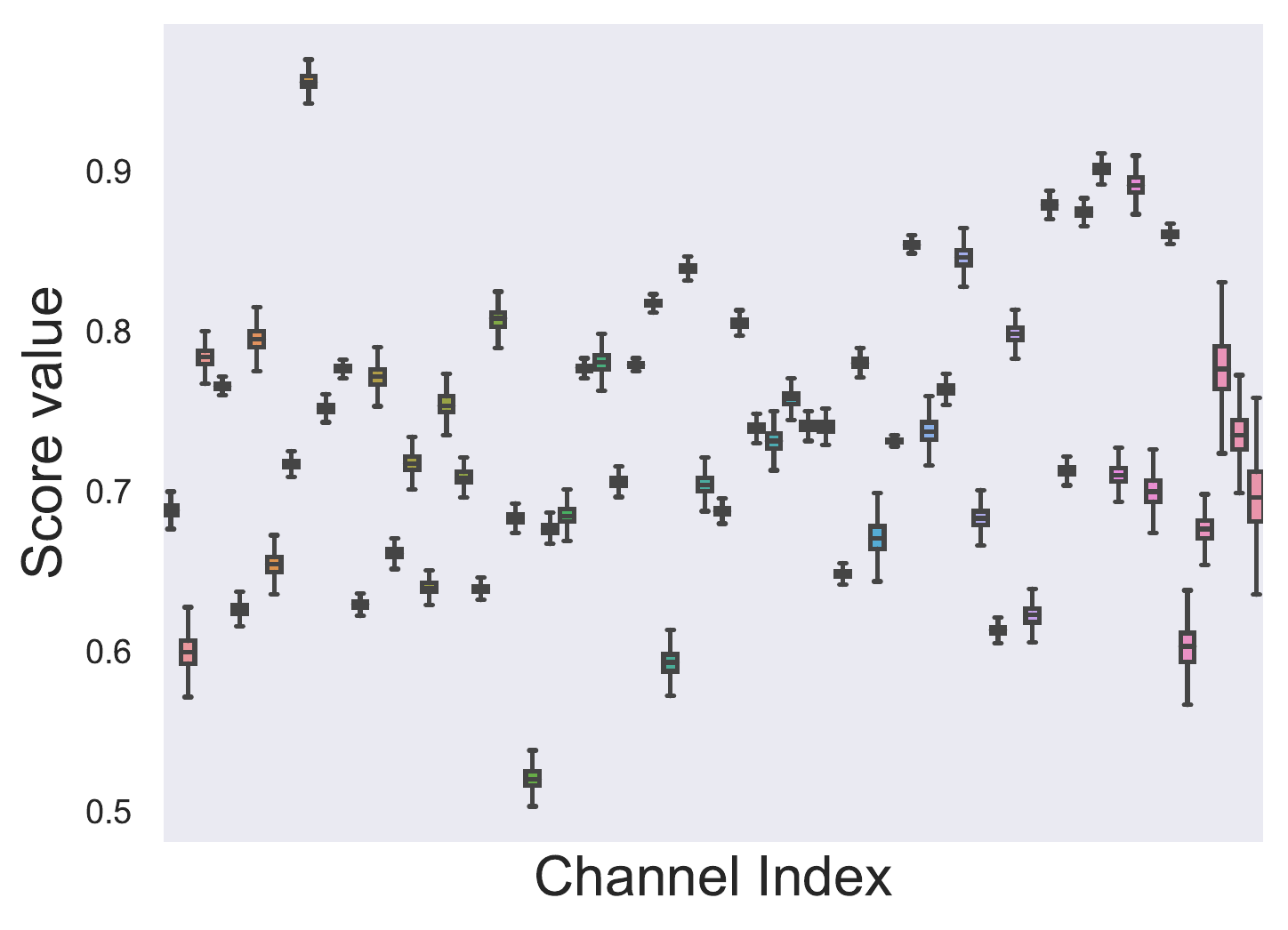}}}%
    \qquad
    \vskip -0.6em
    \caption{The values of learned smoothing scores in the first layer of GFGN-pair.}%
    \label{fig:scores_pair}

\end{figure*}

We provide additional figures for visualization of smoothing scores as shown in Figure~\ref{fig:scores},~\ref{fig:scores_neighbor},~\ref{fig:scores_pair}. For GFGN-neighbor and  GFGN-pair, we choose the scores from its first layer and for each dimension, we group the values from the same neighborhood and plot them in the form of box plot. Similarly, we plot the pairwise values of scores from the first layer of GFGN-pair on each dimension. From the results, we find that similar trends in GFGN-graph can be observed as  in Section~\ref{sec:score_analysis}. On the other hand, the score values learned by GFGN-neighbor and GFGN-pair also vary in a range. Furthermore, since the height of every box in the box plot indicates the variance of score values for each dimension, the various heights of boxes suggest that both GFGN-neighbor and GFGN-pair can also learn adaptive smoothing scores with respect to nodes and pairs. We also note that the variance of the scores in each dimension, represented by the height of the box, shows different trends on different datasets: on datasets with user interests as node features, i.e., BlogCatalog and Flickr, the variance is smaller than that on datasets with bag-of-words features, such as  Texas, Cora and Citeseer. 

\begin{figure*}[h]%
\vskip -0.2em
    \centering
    \subfloat[BlogCatalog]
        {{\includegraphics[width=0.25\linewidth]{images/scores/BlogCatalog.pdf}}}%
    \subfloat[Flickr]{{\includegraphics[width=0.25\linewidth]{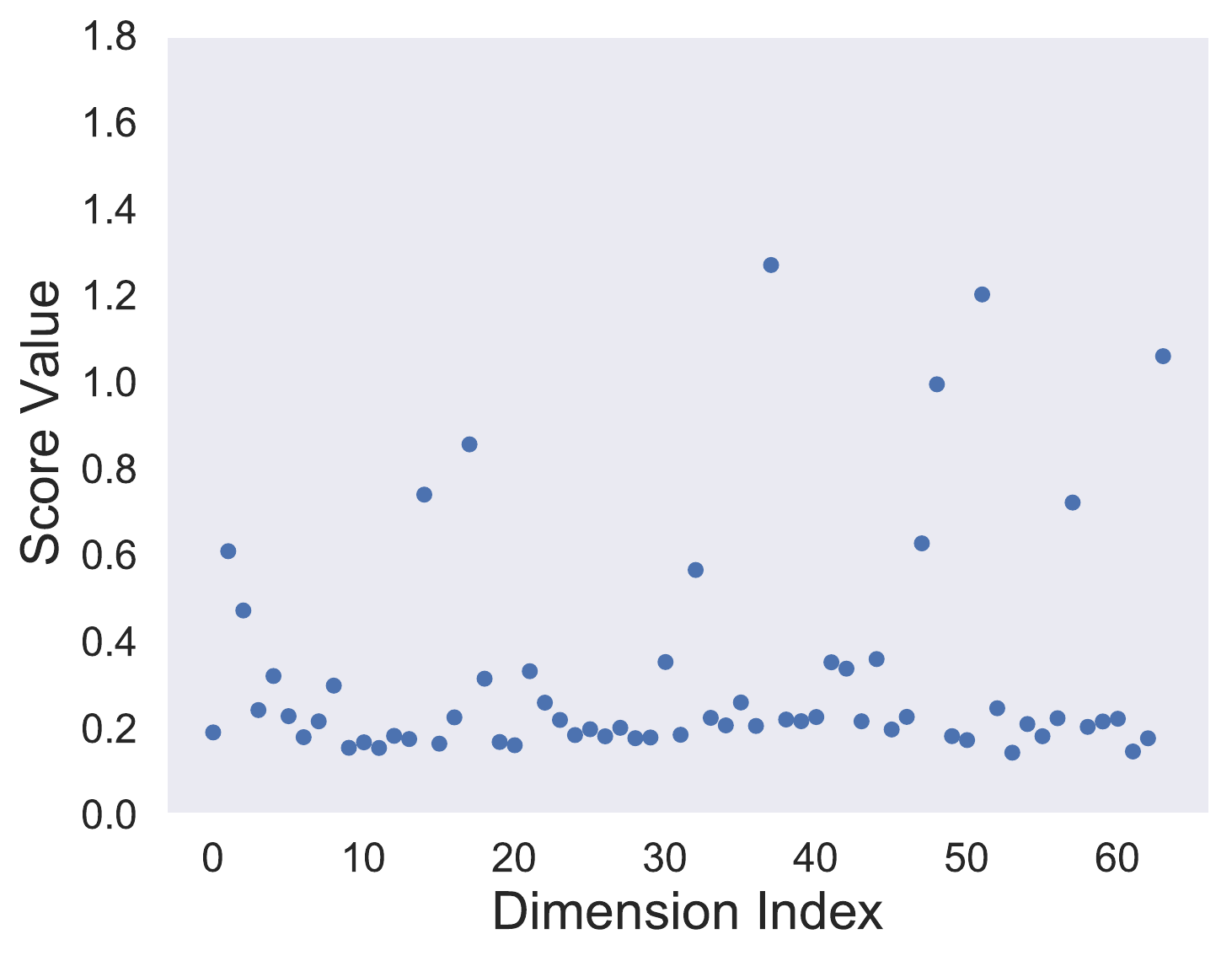} }}%
    \subfloat[Texas]{{\includegraphics[width=0.25\linewidth]{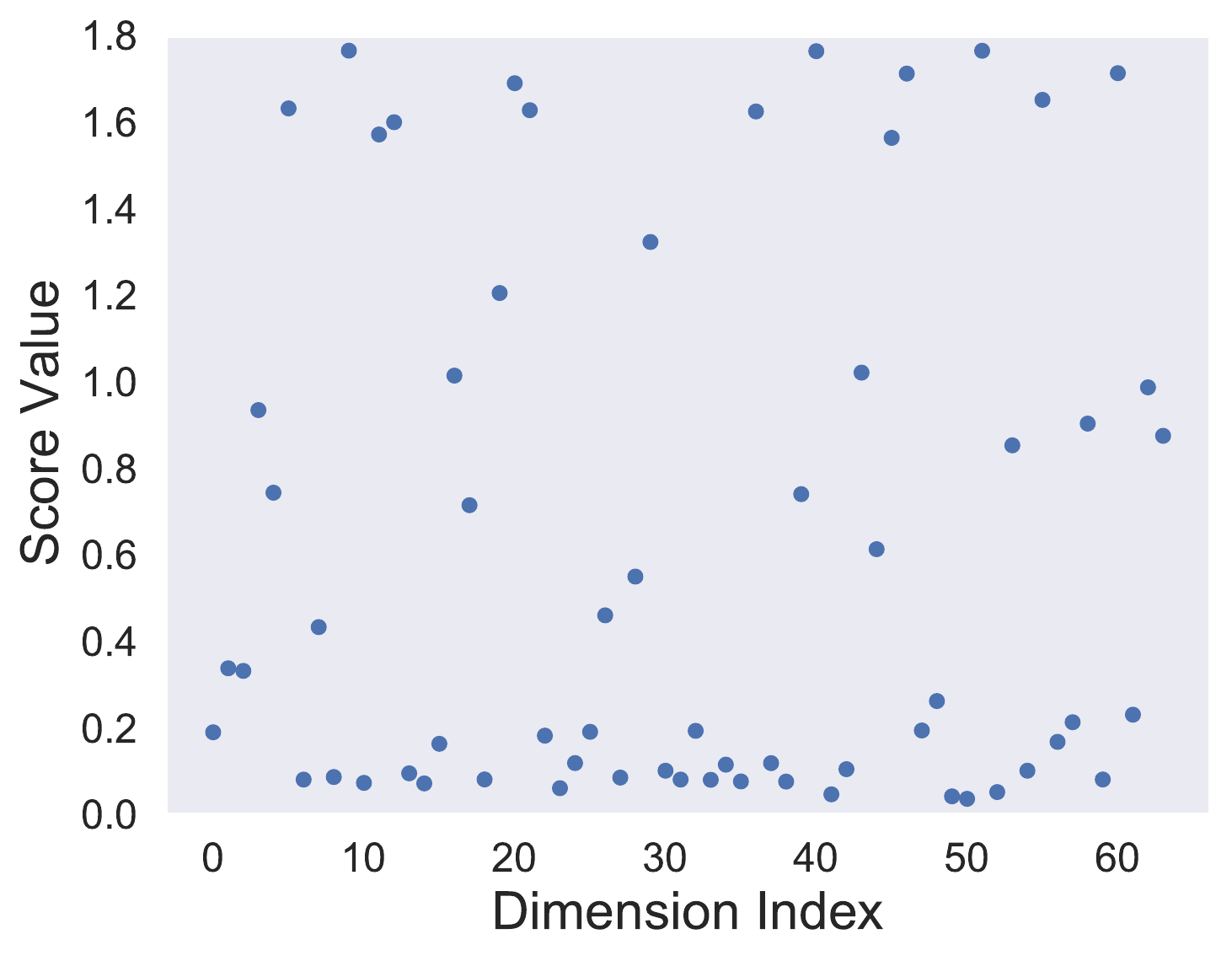} }}%
    \subfloat[Wisconsin]
        {{\includegraphics[width=0.25\linewidth]{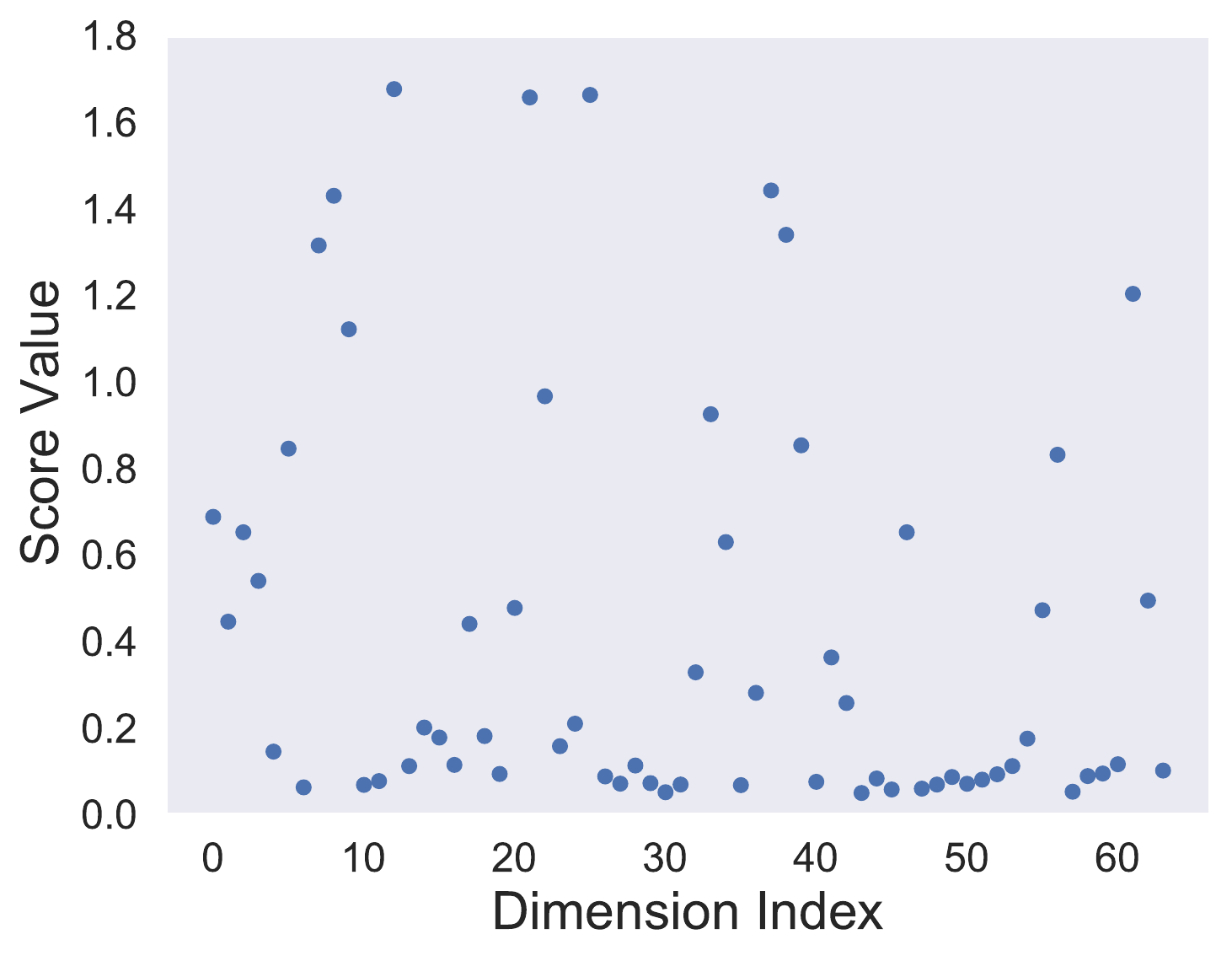}}}%
    \qquad
    \subfloat[Cornell]{{\includegraphics[width=0.25\linewidth]{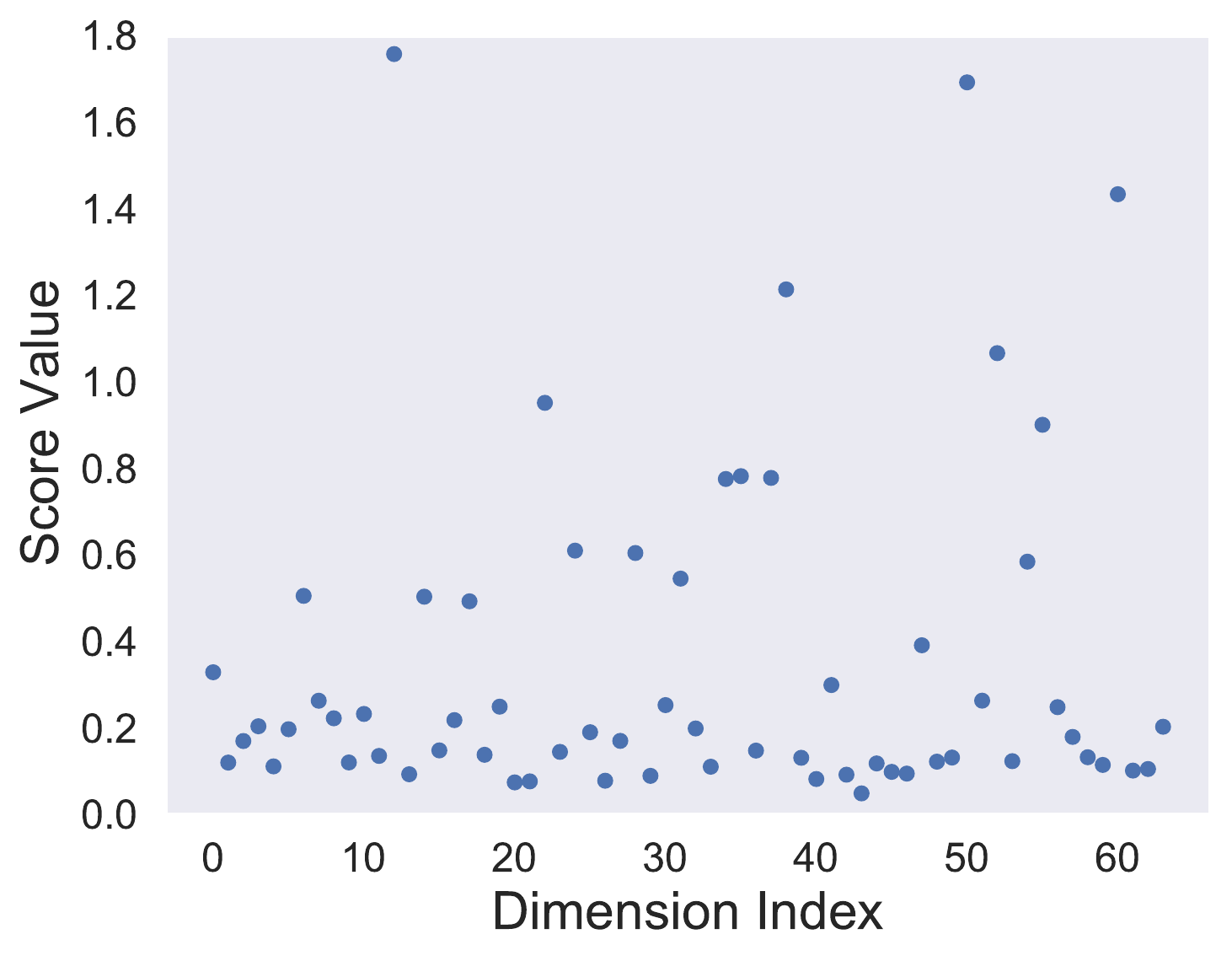}}}%
    \subfloat[Cora]{{\includegraphics[width=0.25\linewidth]{images/scores/cora.pdf} }}%
    \subfloat[Citeseer]{{\includegraphics[width=0.25\linewidth]{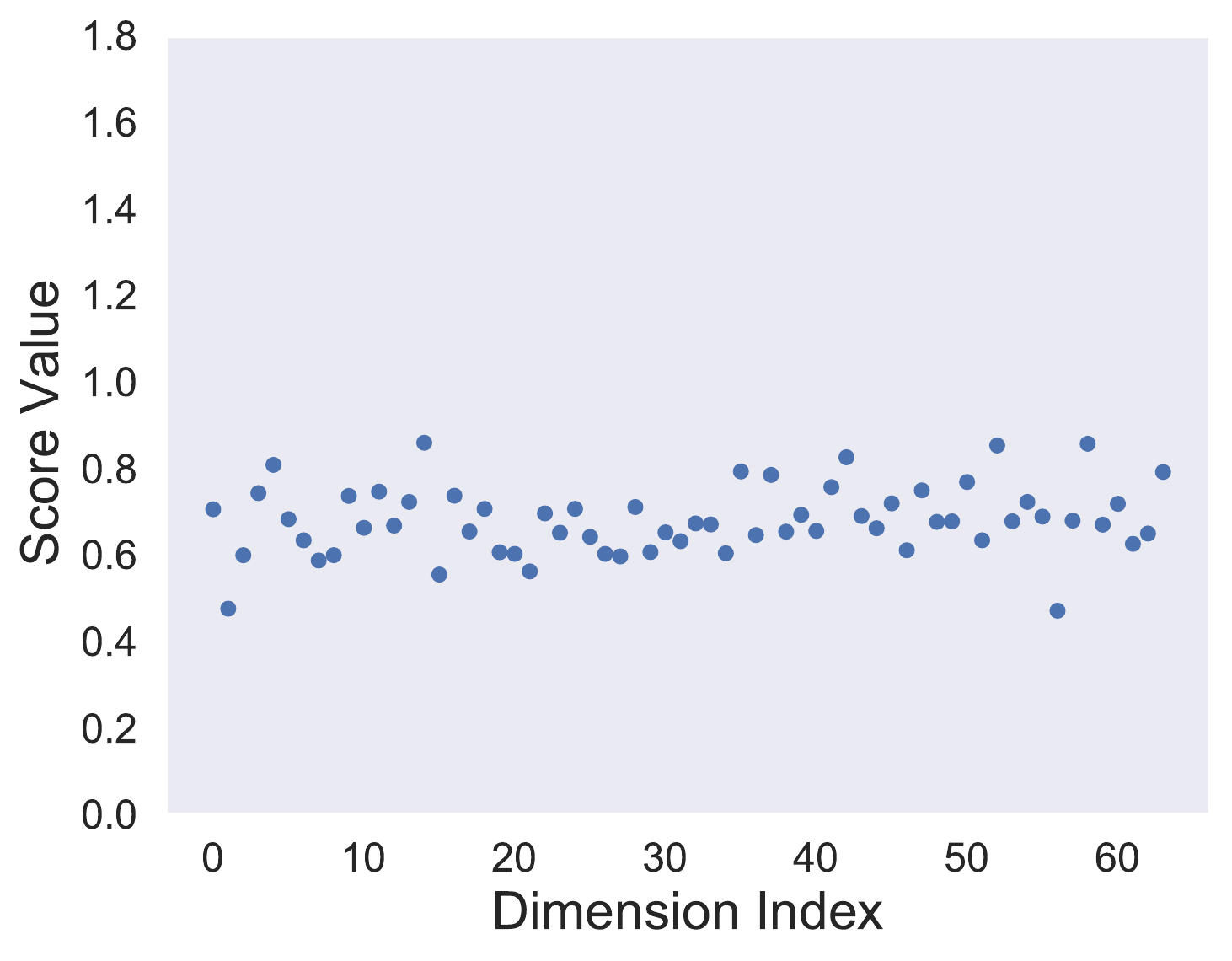}}}%
    \subfloat[Pubmed]{{\includegraphics[width=0.25\linewidth]{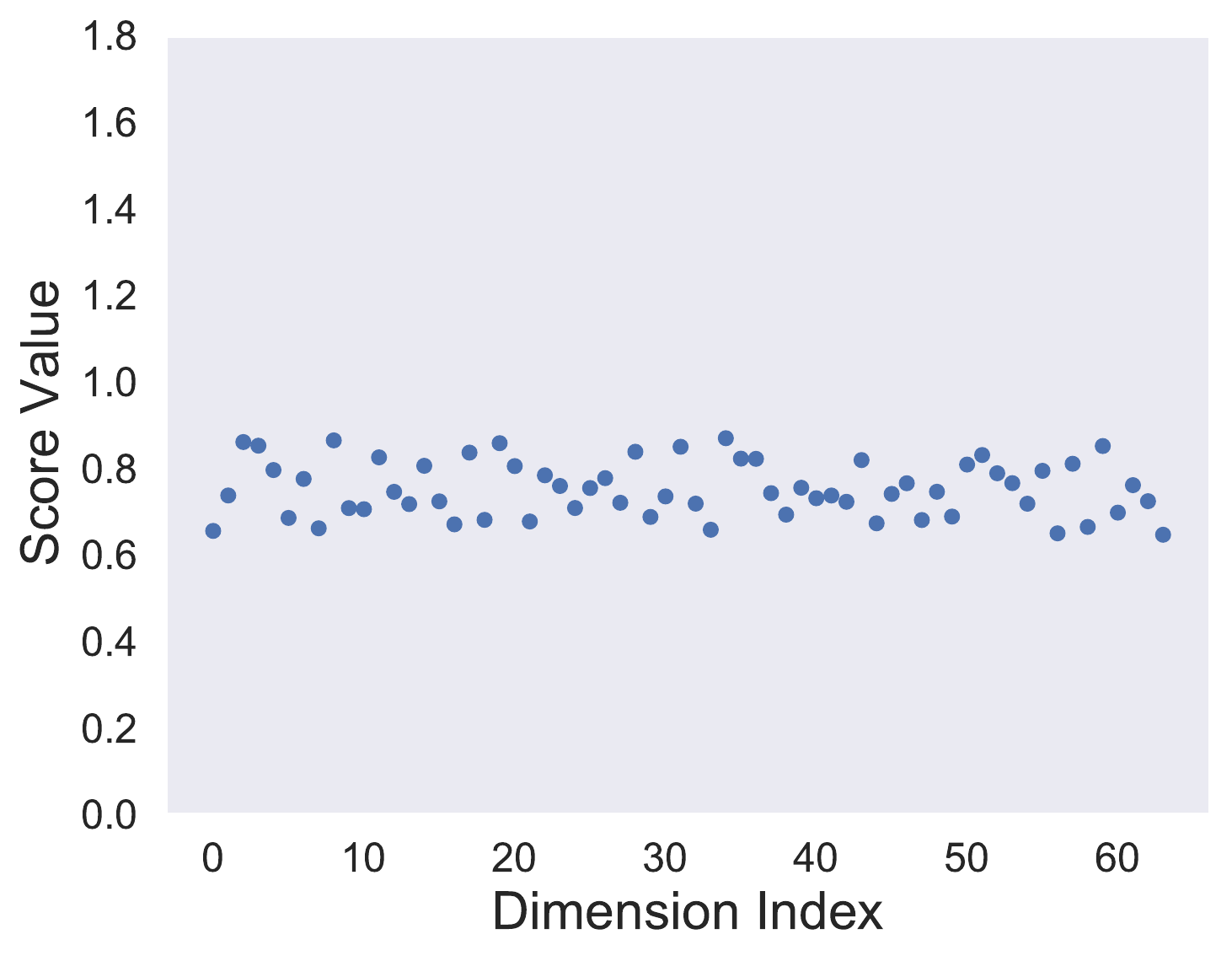} }}%
    \qquad
    \vskip -0.6em
    \caption{The values of learned smoothing scores in the first layer of GFGN-graph.}%
    \label{fig:scores}
    \vskip -1em
\end{figure*}

\end{document}